\documentclass[bsc,frontabs,singlespacing,parskip,deptreport]{infthesis}

\usepackage{multirow}
\usepackage{graphicx}
\usepackage{amssymb}
\usepackage{url}

\begin{document}
\begin{preliminary}

\title{Assessing Dataset Bias in Computer Vision}

\author{Athiya Deviyani}

\course{Artificial Intelligence and Computer Science}

\project{4th Year Project Report}

\date{\today}

\abstract{
A biased dataset is a dataset that generally has attributes with an uneven class distribution. These biases have the tendency to propagate to the models that train on them, often leading to a poor performance in the minority class. With the increasing presence of Artificial Intelligence-based systems in our society, an unfair decision has the potential to be more than just a nuisance. In truth, with systems such as facial recognition for criminal identification and the AI-based recruiting pipeline, a prediction made by a biased model can change a person’s life forever.

Unfortunately, fixing the bias problem is not as straightforward as collecting additional data, due to a number of challenges and limitations. In this project, we will explore the extent to which various data augmentation methods alleviate intrinsic biases within the dataset. We will apply several augmentation techniques on a sample of the UTKFace dataset, such as undersampling, geometric transformations, variational autoencoders (VAEs), and generative adversarial networks (GANs). We then trained a classifier for each of the augmented datasets and evaluated their performance on the native test set and on external facial recognition datasets. We have also compared their performance to the state-of-the-art attribute classifier trained on the FairFace dataset. 

Through experimentation, we were able to find that training the model on StarGAN-generated images led to the best overall performance. We also found that training on geometrically transformed images lead to a similar performance with a much quicker training time. Additionally, the best performing models also exhibit a uniform performance across the classes within each attribute. This signifies that the model was also able to mitigate the biases present in the baseline model that was trained on the original training set. Finally, we were able to show that our model has a better overall performance and consistency on age and ethnicity classification on multiple datasets when compared with the FairFace model. Our final model has an accuracy on the UTKFace test set of 91.75\%, 91.30\%, and 87.20\% for the gender, age, and ethnicity attribute respectively, with a standard deviation of less than 0.1 between the accuracies of the classes of each attribute.

}

\maketitle

\section*{Acknowledgements}
I would like to express my special thanks of gratitude to my supervisor Dr. Ajitha Rajan for her invaluable guidance throughout the completion of this project, and for writing dozens of recommendation letters to accompany my graduate school applications. I would also like to thank Dr. Hakan Bilen for his computer vision expertise.

Secondly, I would also like to thank my parents for their endless love and support, and my cats - Bobby, Tobby, Bonbon, Milo, Mochi, Lilo, Stitch, Kitty, Minnie and Minnie’s three newborn kittens (that we have yet to name) - for letting me pick them up and cuddle them every time work gets a little bit too stressful. The true silver lining of this pandemic is being able to spend more time with family. 

Finally, to all my friends I’ve made throughout my university career, thank you for being my home away from home. Our late nights at Appleton Tower will forever be in my heart.

\tableofcontents
\end{preliminary}

\chapter{Introduction}

In the past decade, we have observed the huge success of machine learning applications, from image recognition to online advertising. Along with the rapid increase in the uses of these applications, reports on biased performance have also increased: from voice assistants not recognizing minority accents, to automated employment screening systems rejecting applicants with ‘feminine’ names. A common source of biased model performance is the intrinsic biases within the dataset that was used to train these machine learning models. It is a well known fact that a majority of popular datasets, particularly the ones used in the field of computer vision, have an imbalanced class distribution. This often results in the model performing poorly on the minority class, i.e. the class with the fewest number of instances. 

Unfortunately, attempting to create a dataset that is perfectly balanced amongst all classes is not a simple task due to the numerous limitations, namely cost and availability. In this project, we will take an imbalanced sample from the UTKFace dataset and perform augmentation on the sample images in an attempt to have a dataset that is balanced along the gender, age, and ethnicity attributes. We have experimented with four popular augmentation techniques, namely undersampling, and image generation through geometric transformations, variational autoencoders (VAEs), and generative adversarial networks (GANs). We then train a state-of-the-art attribute classifier on the augmented datasets and compare their performance against a model that was trained on the original imbalanced dataset. We have also tested the performance of the models on external facial recognition datasets, LFWA+ and CelebA. Finally, we compared our best model to the state-of-the-art attribute classifier that was trained on the FairFace dataset.

From our experiments, we were able to get an accuracy on the UTKFace test set of 91.75\%, 91.30\%, and 87.20\% for the gender, age, and ethnicity attribute respectively. We were also able to obtain a standard deviation of less than 0.1 between the classes within each attribute. The model that performs best in most cases is trained on GAN-generated images. Although, it is important to note that training on the geometrically transformed dataset performs almost as well, with a much shorter training time. Furthermore, we evaluated the performance of the models on the LFWA+ and CelebA dataset, and found that even though there is a drop in overall performance, the drop in the models trained on the augmented datasets is not as much as the one in the baseline model.  From this, we were able to deduce that a balanced dataset obtained through data augmentation improves the overall performance and generalizability of the model. It also alleviates the biases that were present in the baseline model.

Additionally, our best models consistently outperform the FairFace model in age and ethnicity classification, however still lag behind in gender classification. Our models also exhibit a more uniform performance across the classes within each attribute. This trend is also observed when running the classifiers on the external datasets.

\section{Motivation}

Automated systems employing the use of Artificial Intelligence are increasingly used in a multitude of applications in our society, from novelties such as adding filters on social media cameras to making more serious decisions such as recruiting. It is important to note that even though we have not reached a point where these machines make decisions for us, these systems are being more and more commonly used in the decision-making pipeline. In the criminal forensics field, for example, an intelligent system has yet to be used to determine the length of a person’s sentence. However, facial recognition systems are used to identify suspects earlier in the process. It is difficult to design a facial recognition system to 100\% accuracy due to the limitations of 21st century technology, therefore one misclassification can actually have a huge impact in the life of an innocent individual.

The regulation of systems involving Artificial Intelligence is inherently difficult, as it is scientifically complex to define a fair and unbiased system. These systems employ various machine learning models that were trained with labeled data on datasets that might contain a variety of societal biases that are propagated to the algorithms. These algorithms will then in result learn discriminating features which are biased towards certain groups, particularly minorities. 

One popular example is uncovered by Bolukbasi et al. \cite{bolukbasi} where they showed that word embedding algorithms, even the state-of-the-art ones such as Word2Vec \cite{mikolov}, are prone to societal gender biases. When the algorithm is presented with a word analogy task, “man is to computer programmer as woman is to \textit{x}”, it outputs “homemaker”. This shows that the system is propagating already existing societal biases and stereotypes. In another paper, Buolamwini and Gebru \cite{buolamwini} found that several facial recognition and gender classification systems from recognizable entities such as IBM and Microsoft are biased towards people with a lighter skin tone, which happens to be the majority class in the dataset. Both accounts deduce that the main culprit for these discriminatory performances lies within the intrinsic biases in the underlying dataset.

With automated systems being continuously integrated as a functional entity within our society, biases within these systems are getting more than just raised eyebrows. Over the past couple of years, the machine learning fairness field has been getting the attention it deserves, leading to researchers prioritizing creating fairer algorithms and successfully benchmarked discrimination in various contexts \cite{kilbertus}\cite{hardt}. However, only very few of these works are within the field of computer vision.

Only recently, a study was surfaced by Martim Brandão \cite{brandao} that shows that there exists an age and gender bias within state-of-the-art pedestrian detection algorithms, which presents a variety of ethical implications. To evaluate the algorithm for biases, the author had to hand-label images in the INRIA dataset manually, which is a very slow and error-prone method. In healthcare applications, biased samples in medical classification systems can result in treatments that do not work well for minority segments of the population, with an impact similar to the well documented detrimental effects of biased clinical trials presented by Melloni et al. \cite{melloni} and Popejoy and Fullerton \cite{popejoy}. Even after Esteva et al. \cite{esteva} successfully showed in 2017 that simple CNNs can be used to identify melanoma with accuracies as high as experienced medical professionals, it is impossible to gauge just how fair the skin cancer detection system is on different groups of people without additional information about the skin color.

The main difficulty of auditing existing computer vision systems is because of the lack of labels for accompanying so-called soft facial biometric attributes such as gender, age, and ethnicity in most recognition datasets. Obtaining these labels are not as straightforward as one might think, as there are privacy issues concerned when obtaining protected attributes of a person. Therefore, it is preferable to solve annotation issues at the algorithm level rather than dataset collection level. 

Regardless, these particular attributes themselves have attracted the attention of the pattern recognition community, which is hugely contributed by the amount of possible applications in retail and video surveillance. Of course, designing fair and reliable algorithms is challenging in real-world applications, however this is not stopping a lot of researchers from attempting to build such algorithms for face recognition and verification \cite{ding}, expression recognition \cite{li}, gender recognition \cite{greco}, and age estimation \cite{carletti}. 

\section{Project Goals}

The primary goal of this project is to investigate the extent to which the various dataset augmentation techniques are able to mitigate biases within an imbalanced dataset, particularly within the gender, age, and ethnicity attributes. The datasets that will be used in the project are primarily facial recognition datasets, however we would expect that our findings can be extended and applied to various computer vision datasets that are used for other applications such as object detection. Additionally, we are also interested in investigating the generalizability of the models trained on the different augmented datasets by performing cross-dataset generalization on other facial recognition datasets.

As mentioned in the previous section, the observable lag in measuring the accountability of computer vision systems with respect to its fairness is caused by the unavailability of protected attribute labels on most vision datasets. To this day, there are only very few publicly-available datasets with all three labels - gender, age, and ethnicity - available, such as UTKFace and LFWA+. This poses a challenge to those who wish to evaluate how the performance of a model varies throughout the different classes in each attribute. 

Since collecting these attributes is an arduous and possibly costly task, most ethical AI researchers have resorted to manually annotating available datasets. In response to the lack of widely-used and publicly-available automatic attribute annotators for facial recognition datasets, this project will also aim to train a state-of-the-art deep neural network on the augmented datasets and obtain a model that classifies each attribute from a given image with consistent performance across the classes. To further gauge the feasibility of using our final model as a potentially novel automatic attribute annotator, we will evaluate the performance of the model on a variety of facial recognition datasets. 

We will also compare the performance of our final model on an attribute classifier trained on a balanced dataset called FairFace \cite{fairface}, which claims to have high classification accuracies on both the majority and minority classes. The primary difference between the FairFace dataset and the resulting dataset in this project is that FairFace is balanced by collecting data \textit{externally}, while our dataset is balanced through generating images \textit{internally} through various augmentation techniques. The FairFace dataset will be discussed more in-depth in the following chapters.

\section{Summary of Results}

We have trained an attribute classifier on various augmented versions of the UTKFace dataset in an attempt to reduce the imbalance in the class distributions within the gender, age, and ethnicity attributes. We have employed several augmentation techniques such as undersampling, geometric transformations, variational autoencoders (VAEs) and generative adversarial networks (GANs). We aim to produce an unbiased model with a high overall accuracy and F1-score. An unbiased model typically has a uniform performance across the different classes, shown by a low standard deviation between the accuracies and F1-scores of the classes in each attribute.

After thorough experimentation and evaluation, we have observed that the augmentation technique to obtain the best performing model highly depends on the attribute that we are trying to balance. For the gender and ethnicity attribute, we have found that training a model on StarGAN-generated images yields the best performance and uniformity. However, for the age attribute, the best performance is obtained through training on the geometrically transformed images. The summary of performance of the best models for each attribute is shown in table 1.1. A more detailed performance report on various datasets and comparison with the state-of-the-art attribute classifier will be presented in chapter 4.

\begin{table}[h]
\centering
\resizebox{\textwidth}{!}{%
\begin{tabular}{|l|c|c|c|c|c|c|}
\hline
\multirow{2}{*}{Feature} & \multicolumn{2}{c|}{UTKFace} & \multicolumn{2}{c|}{LFWA+} & \multicolumn{2}{c|}{CelebA} \\ \cline{2-7} 
                    & Accuracy & Std. Dev. & Accuracy & Std. Dev. & Accuracy & Std. Dev. \\ \hline
Gender (StarGAN)    & 0.917    & 0.027     & 0.910    & 0.008     & 0.833    & 0.018     \\ \hline
Age (Geometric)     & 0.913    & 0.023     & 0.822    & 0.069     & 0.745    & 0.030     \\ \hline
Ethnicity (StarGAN) & 0.872    & 0.018     & 0.741    & 0.143     & N/A      & N/A       \\ \hline
\end{tabular}%
}
\caption{Performance of best models on the different facial recognition datasets}
\label{tab:my-table}
\end{table}

An important point to note here is that the models trained on a geometrically transformed dataset also yields results that are similarly high-performing and uniform as the models trained on StarGAN-generated images. The main discrepancy here is the training time - while training and generating images from the StarGAN took around 40-50 hours, generating images through geometric transformations require no additional training and only took several minutes. Therefore, we can conclude that even though data augmentation using StarGAN wins by a few accuracy points, using a traditional augmentation technique is the best compromise with respect to accuracy and efficiency.

\section{Structure of the Report}

Chapter 1 is the introductory chapter, where we will present the motivation behind the project and the respective goals and contribution. In chapter 2, we will explore and discuss previous work and state-of-the-art solutions to mitigating dataset biases in computer vision datasets and identify their potential usability for our project. After exploring the limitations of existing methods, we will then introduce our own set of techniques that will be used throughout the project in chapter 3. In chapter 4, these methods will then be evaluated through a series of image classification experiments, and we will compare and justify the results in chapter 5. Finally in the last chapter, we will identify points of improvement for future work and provide a general summary outlining the key takeaways of the project.


\chapter{Background and Literature Review}

\section{Types of Biases}

The unfortunate fact is almost all big datasets generated by systems powered by Machine Learning models are known to be biased. In the past decade, we have observed the skyrocketing success of machine learning applications, from online advertising to image recognition, and have been adopted to daily life applications, from phones with built-in voice assistants to smart homes. As these devices become more common in society, there has been a disturbing rise in reports of gender, race, and other types of bias in these systems - from ad ranking systems being accused of racial and gender profiling \cite{sweeney} to Amazon having to shut down a model that scores candidates for employments due to its tendency to penalize women \cite{amazon} - and oftentimes, these biases can be traced back to the dataset being used.

Due to the visual nature of computer vision datasets, it is not surprising that predefined image collections present easily recognizable biases. These primary causes have been pointed out and comprehensively described by Torralba et al. \cite{torralba} as the following: 

\begin{itemize}
\item
\textbf{Selection bias} is the tendency of datasets preferring kinds of images, such as street scenes, nature scenes, or images retrieved via Internet keyword searches. Selection bias occurs when a dataset does not reflect the realities of the environment in which a model will run. For example, in a facial recognition task, the model is trained primarily on images of white men. This model would have a considerably lower level of accuracy when tested against the faces of women and people of different ethnicities. 
\item
\textbf{Capture bias} is related to how the images are acquired both in terms of the used device and of the collector preferences for point of view, lighting conditions, object positioning, angles, etc. It also takes into account the fact that photographers tend to take pictures of objects in similar ways.
\item
\textbf{Category/label bias} comes from the fact that semantic categories are often poorly defined, and different labelers may assign different labels of the same type of object, e.g. "grass" vs. "lawn" and "painting" vs. "picture". Sometimes, the converse is true, i.e. the same label can also be assigned to visually different images.
\item
\textbf{Negative set bias} defines what the dataset considers to be "the rest of the world". If we focus only on the classes shared by the different datasets, the 'rest of the world' will be defined differently depending on the collection.
\end{itemize}

The presence of any of the above may cause the object recognition dataset to be not fully representative of the domain it is trying to represent, which in our case happens to be the 'real' world, and thus being \textit{biased}. A biased dataset could produce classifiers that are overconfident and not very discriminative, and it might, in the extremes, also cause several ethical and legal issues. The following section will discuss previous work that has attempted to perform critical evaluation and mitigate these biases.

\section{Previous Work}

This section introduces some related studies from the past decade and provides a brief review of the methods attempted previously. This section is divided into two parts: the first part will discuss previous methods used to evaluate biases within object recognition datasets, and the second part will discuss previous studies that are concerned with alleviating biases within object recognition datasets. 

\subsection{Evaluation}

The growth of the object recognition field can be attributed to the availability of vast datasets. Not only do they provide a large amount of training data, but they also provide means of measuring and comparing the performance of competing algorithms. However, Torralba and Efros \cite{torralba} expressed their concern about how research surrounding object recognition puts too much focus on breathing the latest benchmark numbers on the latest dataset to the extent that they might have lost sight of the original purpose of the field.

They conducted a study to compare popular object recognition datasets and evaluate them based on several criteria. The paper is also the first to conduct cross-dataset generalization to evaluate biases within the datasets, and more on the method will be discussed further below. This paper served as a wake-up call to the computer vision field to address the dataset bias issues as its applications are growing just as vast as the field. The methods used in the paper, and later on in various studies in the following years, to take stock of the current state of recognition datasets will be also be discussed.

\subsubsection{Name That Dataset!}

The goal of the 'game' called \textit{Name That Dataset!} is to guess which images came from which dataset. Theoretically, this would be a challenging task considering that the datasets contain thousandst to millions of images and that they were collected with the goal of being as varied as possible, aiming to sample the visual world 'in the wild'. 

This evaluation method looks at the most discriminable images within each dataset, i.e. the images placed furthest from the SVM's decision boundary. The opposite method is also possible: for a given dataset, look at the images placed closest to the decision boundary separating it for another dataset. This shows how one dataset can 'impersonate' a different dataset.

From this, they have found that, despite the best efforts of the creators of the datasets, they appear to have a strong built-in bias. However, most of the bias can be attributed to the different goals of the different datasets. They have also concluded that even if the capture biases are controlled by isolating specific objects of interest, the biases will still be present in one form or the other.

\subsubsection{Cross-Dataset Generalization}

As mentioned previously, there has not been any paper demonstrating cross-dataset generalization to assess an object recognition dataset. Theoretically, this task should be easy if the datasets were truly representative of the real world, and would give access to much more labelled data. 

Previous methods discussed in various papers \cite{lixin}\cite{feifei}\cite{yang} involve transferring a model learned on one dataset into another. However, Torralba and Efros \cite{torralba} points out an interesting issue: these methods consider the target dataset as a different domain, even though the datasets are trying to represent the same domain - our visual world!

Cross-dataset generalization aims to answer the following question: how well does a typical object detector trained on one dataset generalize when tested on a representative set of other datasets, compared with its performances on the ‘native’ test set? They chose two classes that were common among all datasets, ‘car’ and ‘person’, and performed detection and classification tasks. The evaluation results show that there is a big performance drop, and that there is little generalization that happens beyond the given dataset.  

From this, they concluded that some popular vision datasets, like Caltech-101 \cite{caltechdataset}, are extremely biased and supported the idea that they should have been ‘retired’ long ago. In addition to that, they also emphasized that this issue should be put at the forefront of object recognition research if our goal is to build algorithms that can understand the visual world. 

\subsection{Dealing with Dataset Bias}

\subsubsection{Domain Adaptation and Transfer Learning}

Domain adaptation aims at solving the learning problem on a target domain (data from real scenarios) by exploiting information from a source domain (data used to train the model) when both the domains and the corresponding tasks are not the same. Transfer learning focuses on the possibility to pass useful knowledge from a source task to a target task with different label sets when the corresponding domains are not the same but the marginal distributions of data are related. When used together, they are able to address the problem of a mismatch between the joining distribution of inputs between source and target domains, also known as the domain shift \cite{candela}. One of the first studies that proposed the idea of domain adaptation for object recognition by Saenko et al. \cite{saenko} involve the idea of learning a regularized transformation using information-theoretic metric learning that maps data in the source domain to the target domain. A later study by Kulis et al. \cite{kulis} generalizes this further to handle asymmetric transformations in which feature dimensionality in source and target domain can be different.

A study by Gopalan et al. \cite{gopalan} addresses the issue in \cite{saenko} and \cite{kulis} that requires labeled data from target domain by proposing a domain adaptation technique for an unsupervised setting where data from target domain is unlabeled. This method obtains domain shift by generating intermediate subspaces between the source and target domain, and then projecting both domains onto the subspaces for recognition.

\subsubsection{Mathematical Frameworks for Multi-Task Learning}

Multi-Task Learning aims at learning jointly over \textit{N} available sets, leading to a symmetric share of information. Evgeniou and Pontil \cite{evgeniou} and Ben-David and Schuller \cite{ben} have proposed a mathematical framework for multi-task learning where solutions to multiple tasks are tied through a common weight vector. This common weight vector is used to share information among tasks but is not constrained to perform well on any task on its own. 

Although similar to \cite{evgeniou} and \cite{ben}, the method proposed by Khosla et al. \cite{khosla} differs by the fact that their goal is to learn a common weight vector that can be used independently and is expected to perform well on a new dataset. Their model, which is based on a discriminative framework, is also novel in the way that it provides a first step to building models that explicitly include dataset bias in the mathematical formulation with the goal of mitigating its effect. Under the assumption that the features used are common for all images from all datasets and that bias between datasets can be identified in feature space (features are rich enough to capture the bias in the images), the discriminative framework will jointly learn a weight vector corresponding to the visual world object model, and a set of bias vectors for each dataset, that when combined with the visual world weights result in an object model specific to the dataset. They formulated the problem in a max-margin learning (SVM) framework similar to the one proposed by Evgeniou and Pontil \cite{evgeniou}. 

Khosla et al. \cite{khosla} evaluated their model by performing two tasks: (1) object classification on seen and unseen datasets and (2) object detection on unseen datasets, and performed in-data and cross-dataset generalization to evaluate their algorithm performance against an SVM baseline. The results prove their algorithm successful as it constantly outperforms the SVM at all occasions, and thus showing that their framework is effective at reducing the effects of bias in both classification and detection tasks. As they’ve compared their model to the common weight vector from \cite{ben}, it made sense to use an SVM as a baseline, however since most object detection and classification tasks today are multi-class problems, a neural network model would be a more reasonable baseline.

\subsubsection{Multi-Task Unaligned Shared Knowledge Transform (MUST)}

The rather disappointing cross-dataset generalization results shown by Torralba and Efros \cite{torralba} led Tommasi et al. \cite{tommasi} to the following hypothesis: a classifier trained on a specific dataset learns a model containing some generic knowledge about the semantic categorical problem, and some specific knowledge about the bias contained into that dataset. From this, they decided to propose an algorithm that focuses on improving cross-dataset generalization performance when trying to mitigate dataset bias. Similar to \cite{torralba}, their method exploits existing visual datasets preserving their multiclass structure and relying on the fact that each of them present specific characteristics, but all together they cover different nuances of the real world.

Their Multi-Task Unaligned Shared Knowledge Transform (MUST) algorithm combines the techniques that have been used so far - domain adaptation, transfer learning, and multi-task learning. It aims to extract general information from all the sources in multi-task fashion to use it when learning on a new target with a general advantage both on the known categories (domain adaptation) and on new ones (transfer learning). The algorithm learns a projection function based on the folk-wisdom principle: pulls objects or data samples together if they are the same type and pushes them apart if they are not. The algorithm decomposes multiple datasets into two orthogonal subspaces - one is specific to each dataset and the other is shared between all of them, then the common information is transferred to help on a new task. 

The MUST algorithm is evaluated through the single-view setting, where the same features were used for each dataset, and a multi-view setting, where different features were used for each dataset. The multi-view setting is particularly useful as it retains dataset-specific characteristics before inferring the shared knowledge in successive iterations. Khosla et al. \cite{khosla} compared MUST with a set of baseline models through a cross-dataset generalization evaluation and found that their algorithm outperformed others for both settings on average. This provides evidence that they have achieved their goal to show that datasets do carry a useful knowledge which is learnable and exploitable regardless of the bias afflicting them, and significantly improving the generalization ability of a learning system. The MUST algorithm also overcomes the class alignment limit of the SVM multiclass models. However, they did perform evaluation on the target dataset they trained on instead of on a completely new unseen dataset, which makes it unlawful to call it cross-dataset generalization as the model was generalizing on samples from the same domain. In addition to that, multi-task learning is particularly useful when each task has few data thus the results from the experiments cannot be generalized to the vast object recognition datasets available today.  

\subsubsection{Image Descriptors}

Later on, the authors of MUST \cite{tommasi}, explored the potential of DeCAF \cite{decaf}, a robust feature representation learned by convolutional neural networks (CNNs), when facing the dataset bias problem. Through this, they aim to answer the question of how we can use available data to generalize new unseen training samples even when training and test collections are different. They used existing debiasing methods and used a less powerful image descriptor, BOWsift, as a baseline for comparison. 

When doing the \textit{Name the Dataset!} test, they found that DECAF has better separation among collections than BOWsift, and that there are high confusion levels to datasets with large number of classes and images per class and low confusion levels for those that are more specific. They also performed the cross-dataset generalization test on two object classes that are shared among multiple datasets: ‘car’ and ‘cow’. They found that non-rigid objects like cows are more challenging to classify compared to rigid objects like cars due to its large in-class variability.

From their comprehensive experiments they performed, they concluded that DeCAF not only does not solve the dataset bias problem in general, but in some cases (both class- and dataset-dependent) they capture specific information that induce worse performance than what obtained with less powerful features like BOWsift. In addition to that, highly descriptive power of the features that determined much of their successes makes the task of learning how to extract general information across different data collection more difficult, and that a simple selection procedure based on self-labelling over the test set leads to a significant increase in performance.

\subsubsection{FairFace}

While the previous methods aim to tackle the dataset bias through improving existing algorithms, Karkkainen et al. decided to approach the problem from another angle. The creators of the FairFace dataset \cite{fairface} were primarily interested in how most face datasets are strongly biased toward Caucasisan faces. They also investigated the impact of an imbalanced dataset to the consistency of the model accuracy and applicability of systems trained on these datasets on non-Caucasian users. To solve this problem, they constructed a novel face image dataset containing over 100,000 images that prioritizes a balanced ethnicity composition, containing images collected from the YFCC-100M Flickr dataset. The labels in the FairFace dataset are race, gender, and age groups. This dataset is notably one of the most complete dataset currently available for ethnicity recognition. A random sample containing several images from the FairFace dataset selected by the authors is shown in figure 2.1. 

\begin{figure}
\centering
\includegraphics[width=0.5\textwidth]{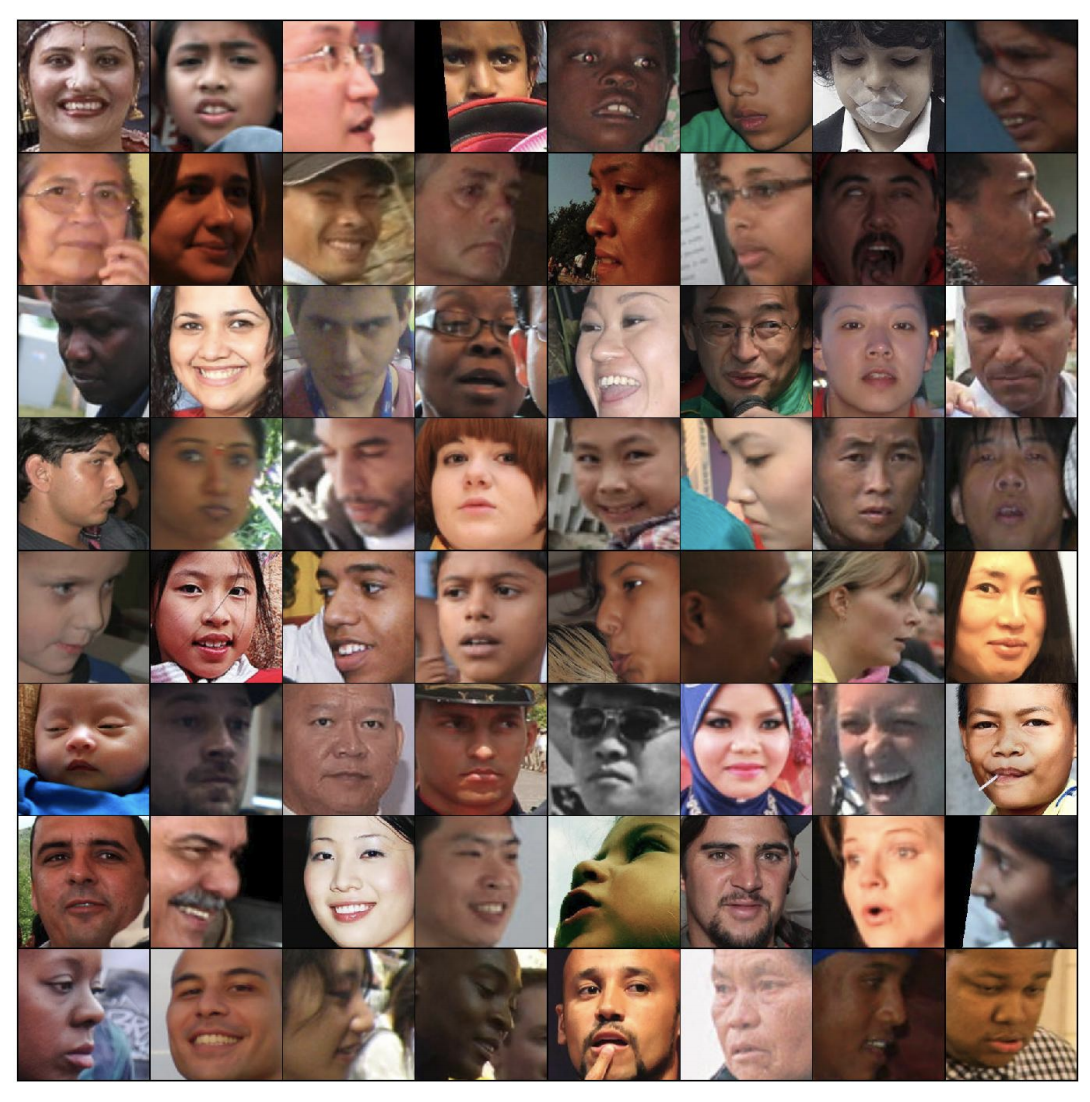}
\caption{Sample images from the FairFace dataset \cite{fairface}}
\end{figure}

Furthermore, the authors of the FairFace dataset have evaluated the gender, age and ethnicity classification performance of a ResNet-34 model trained using different training sets. The experiments involve evaluations on different test sets, in order to investigate the generalization capabilities achieved by the network trained with a specific set. They were able to show that the ResNet-34 that was trained on the FairFace dataset generalizes substantially better than the same model trained on UTKFace \cite{utkface} and LFWA+ \cite{lfw}. This demonstrates the importance of using their balanced FairFace dataset for training an attribute classifier.

Additionally, Greco et al. have benchmarked deep network architectures \cite{greco_benchmark}  namely VGG-Face \cite{vggface}, ResNet-50 \cite{resnet50}, and MobileNet v2 \cite{mobilenet}, for ethnicity recognition. They have trained the models on a variety of facial recognition datasets with ethnicity attributes such as UTKFace, LFWA+, MORPH-II \cite{morph2}, and FERET \cite{feret}, that were collected and labeled in numerous different ways. From a series of experiments, they were able to show that FairFace is the only dataset that was able to provide the network models with generalization capabilities, however the performance on different test sets still show a certain degree of variation.

\subsection{Generative Data Augmentation}

Most imbalances are usually caused by the inability to collect additional data in order to create a dataset with an even class distribution. Therefore, researchers have explored the plausibility of generating new data by augmenting images from existing datasets. Shorten et al. have surveyed and discussed the various state-of-the-art augmentation techniques in their recent work \cite{shorten}. The paper defines data augmentation as generative modeling, which often refers to the practice of creating artificial instances from a dataset such that they retain similar (yet not identical) characteristics to the original set. Amongst all the data augmentation techniques covered in their paper, they have divided them into two general methods: traditional and CNN-based. The methods described below will be more thoroughly discussed in the next chapter.

Traditional augmentation involves performing basic manipulations to the source image. There are a variety of manipulations that can be done, namely flipping, cropping, rotating, shearing, translation, noise injection, and more. Choosing which method to use requires understanding the context of their ‘safety’ of application. The safety of a data augmentation method refers to its likelihood of preserving the label post-transformation. For example, rotations and flips are generally safe on facial recognition datasets, but not for digit recognition tasks as flipping a ‘9’ by 180 degrees will lead to a different digit, ‘6’. 

Another popular augmentation method is the use of autoencoders. They are especially useful for performing feature space augmentations on data. The encoder and decoder network work simultaneously to map images to a low-dimensional vector representation and reconstruct the vectors back into the original image respectively. By extrapolating between the 3 nearest neighbors per sample, DeVries and Taylor \cite{devries} were able to generate new data that are similar but not identical to the input source. It is also possible to do feature space representation by isolating vector representations from a CNN using a CNN-based autoencoder. An improvement from the standard autoencoder is the variational autoencoder \cite{vae_tutorial}. A variational autoencoder is an autoencoder whose encodings distribution is regularised during the training to ensure that it has a continuous latent space that allows us to generate some new data, thus making sure that the generated image looks ‘realistic’ and resembles the input images.

Finally, with the recent growth in deep learning brought forth the possibility of using adversarial training to generate images from an existing dataset. Adversarial training is a framework for using two or more networks with contrasting objectives encoded for their loss functions. A popular generative modeling framework based on the principles of adversarial training is the Generative Adversarial Network, or GAN. First proposed by Ian Goodfellow \cite{ian_gan}, the main idea behind a GAN is a generator network that tries to generate realistic-looking images based on the input that can ‘fool’ a discriminator network to think that the generated image comes from the input. The success factor lies when the discriminator can no longer identify whether a generated image is from the training set or created by the generator network. Since its introduction, a variety of architectures have been proposed, from DCGAN \cite{dcgan}, CycleGAN \cite{cyclegan}, to StarGAN \cite{stargan}. A recent survey conducted by Yi et al. \cite{yi_gan_survey} covers the use of GANs in image reconstruction applications such as CT denoising \cite{gan_ct} and accelerated magnetic resonance imaging \cite{gan_mri}. The survey also covers the use of GAN image synthesis in medical imaging applications such as brain MRI synthesis \cite{gan_mri_synthesis} and lung cancer diagnosis \cite{gan_lung_cancer}.


\chapter{Methodology}

Dataset imbalance is a problem inherent to even state-of-the-art datasets that are used in the majority of our intelligent systems today. The distribution of a dataset can vary from a slight bias to a severe imbalance where an occurrence of a particular class is very scarce. This is troublesome for predictive modeling because most of the machine learning algorithms used for tasks such as classification were designed around the assumption that the dataset is balanced class-wise. As a result, we will have models with poor predictive performance, especially for the minority class. This is problematic as the minority class is often the important class, such as a positive label in a cancer detection dataset. The model would be more sensitive to classification errors for the minority class than the majority.

Due to the availability and cost limitations, it is almost impossible to solve the imbalance problem by simply ‘adding new data’. Therefore, we have to look at techniques that can be done to obtain a balanced dataset from an imbalance source. Following the limitations of the methods from previous work which attempted to alleviate biases in object recognition datasets, we decided to introduce other popular methods that have been employed to mitigate biases and address the imbalance problem in different machine learning fields, such as undersampling and geometric transformations. We will also be introducing the use of Variational Autoencoders (VAEs) and Generative Adversarial Networks (GANs), even though their usage for alleviating biases in facial recognition datasets has yet to be made conventional. Therefore, we are interested to investigate their ability to address the dataset bias problem.

These aforementioned techniques come with their own set of benefits and drawbacks, which is why we must critically evaluate each of them. It is important to note that all the data augmentation methods used throughout this project are introduced and discussed in this chapter, while their detailed technical implementation will be elaborated further in the following chapter.

\section{Data Augmentation Techniques}

\subsection{Undersampling}

Random undersampling \cite{kotsiantis} is a non-heuristic method that aims to balance class distribution through the random elimination of majority class examples. We chose this method instead of oversampling as it has been shown that undersampling outperforms oversampling \cite{fan} and oversampling can lead to overfitting \cite{chawla}. Additionally, oversampling will lead to the minority class having a reduced variance, potentially leading to poorer generalization performance. In undersampling, we keep all instances of the minority class and randomly sample, without replacement, an equal proportion from the majority class. The resulting dataset is then used to train the classifier. This aims to balance out the dataset to overcome the idiosyncrasies of the machine learning algorithm. Random undersampling can also be useful to remove variances within the majority class, and thus the machine learning algorithm will only learn the most prominent features of the class. One obvious drawback of random undersampling is that this method might remove potentially useful features that are unique to a certain class. 

In addition to that, when we train a machine learning classifier, we are essentially teaching the classifier to estimate the probability distribution of the target population. Since that distribution is intentionally left unknown, the classifier will try to estimate the population distribution by using the sample distribution found in the training set. Statistically speaking, as long as the sample is randomly drawn, the sample distribution can be used to estimate the distribution of the target population as it is drawn from the overall population. However, after undersampling the majority class, the overall population distribution no longer corresponds to the target population, and thus the sample cannot be considered random.

Regardless, a variety of undersampling methods have been shown to be effective when it comes to dealing with a dataset with a minority class that is significantly smaller than that of the majority class. Some of the successful applications involve credit card fraud detection \cite{credit_card_fraud}, estimating corporate bankruptcy \cite{corp_bankruptcy}, as well as balancing datasets used in medical applications such as the thyroid and breast cancer dataset covered in \cite{thyroid}. 

\subsection{Geometric Transformations}

Geometric transformation is a form of traditional data augmentation technique that is widely used to balance datasets containing images. Geometric transformation entails cropping, rotating, flipping, zooming, shearing, and more. In addition to its effectiveness in increasing the overall algorithm accuracy, geometric transformation techniques are easily implemented through popular deep learning libraries such as TensorFlow \cite{tensorflow} and PyTorch \cite{pytorch}. 

There are certain cases where the application of geometric transformation on an image dataset needs to be closely monitored. An example dataset that would be sensitive to geometric distortions would be the MNIST dataset \cite{mnist}, where excessive flipping and rotating might lead to inaccurate true labels, such as rotating a number ‘9’ by 180 degrees would turn it into the number ‘6’. However, this paper is primarily concerned with facial recognition datasets, so we do not need to worry too much about these distortions.

One of the most obvious drawbacks of geometrically transforming a dataset is that the resulting images are just very slightly modified versions of the original images, and thus some might consider this method as a moderately ‘smarter’ oversampling method. It can also potentially lead to homogenizing the data if we are planning to generate a large transformed dataset from a relatively small sample. However, even though the images may not look like a ‘new’ set of images to the human eye, subtle spatial discrepancies such as horizontal flipping would be detected by deep learning algorithms, leading them to think that it’s a completely new sample. This is why geometric transformations seem to be effective in alleviating dataset biases and improving the overall algorithm accuracy.

\subsection{Autoencoders}

Autoencoders \cite{ballard} are essentially artificial neural networks that were built to recreate a given input. It takes a set of unlabeled inputs and encodes them, then tries to extract the most valuable information from them. Autoencoders are primarily used for feature extraction, dimensionality reduction, and compression applications.

The general principle behind an autoencoder is to generate a low-dimensional representation of a high-dimensional input, most commonly known as the latent representation. The process of mapping from input to the latent representation is commonly known as representation learning. This is achieved by asking the model to simply recreate the input, while imposing an information bottleneck upon the model so that it is forced to lose a massive amount of information from the original input in the process. In other words, we are forcing the model to learn only invariant features within the input space. From this, we are encouraging the model to encode and retain as much useful information as it passes the bottleneck, resulting in the development of two submodels: the \textbf{encoder} network that takes in an input and converts it into a smaller, dense representation, and the \textbf{decoder} network that converts the dense representation back to the original input. The overall structure of an autoencoder is shown in figure 3.1.

\begin{figure}[h]
\centering
\includegraphics[width=0.5\textwidth]{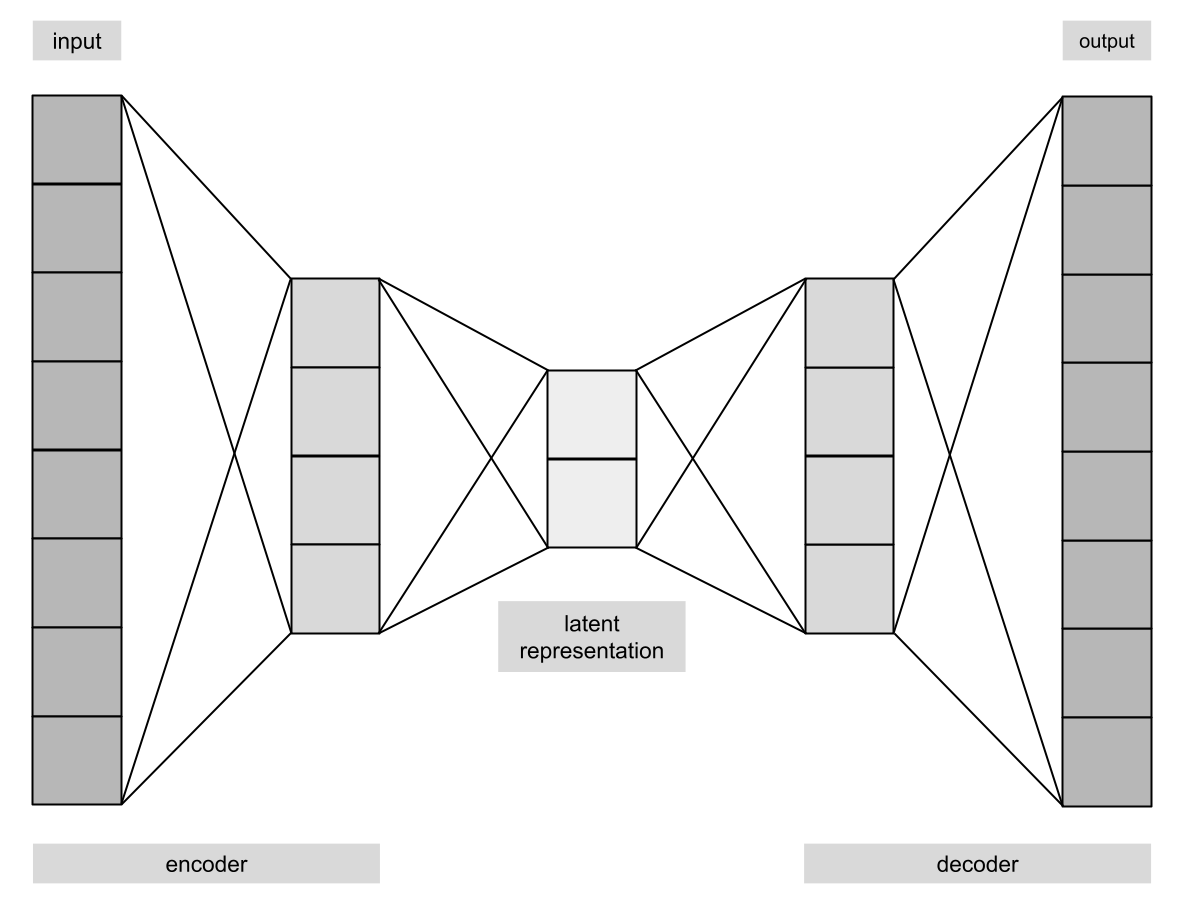}
\caption{A standard autoencoder architecture}
\end{figure}

Similar to other machine learning models, the autoencoder employs a loss function to train the network. The loss function is usually either the mean squared error or cross-entropy between the output and input, which is known as the reconstruction loss. The reconstruction loss penalizes the network for creating outputs that are different from the inputs.

Standard autoencoders are able to generate compact representations and reconstruct their inputs well, however apart from applications such as denoising autoencoders, they have a limited range of applications. Particularly, when we are aiming to use autoencoders as a generative network, we are facing a fundamental problem: the latent space that they convert their inputs to and where their encoded vectors lie might not be continuous. This is completely fine if we simply would like to replicate the same images, however not so much when we want to generate variations of the input image.

A discontinuous space is problematic for standard autoencoders because when attempting to generate a sample from that region, the decoder will generate an unrealistic output as it does not know how to deal with that specific region of the latent space. One of the main reasons for this occurring is that the model has never seen encoded vectors coming from that region of the latent space during training. To solve this problem and to make autoencoders as a useful generative model, Diederik Kingma and Max Welling \cite{kingma} came up with the variational autoencoder.

\subsubsection{Variational Autoencoders}

Variational autoencoders possess a unique property that sets them apart from standard autoencoders: their latent space is continuous, allowing easy random sampling and interpolation. This property is what makes them so useful for generative modeling. This is done by representing the encoding output as two vectors, instead of directly learning the latent representation from the input. The two vectors are the a vector of means \(\mu\), and another vector of standard deviations \(\sigma\). The vectors form the parameters of a vector of random variables, which is where we obtain the sampled encoding to be passed to the decoder. Aligned with their statistical definitions, the mean vector controls where the encoding of an input should be centered around, while the standard deviation vector controls how much from the mean the encoding can vary.

In a variational autoencoder, the decoder is able to decode encodings that slightly vary from the original encodings from the latent space. This is because the decoder is exposed to a range of variations of the encoding of the same input during training. After training, the model is now exposed to a certain degree of local variation by varying the encoding of one sample, resulting in a smooth latent space.

Ideally, we want encodings which are as close as possible to each other while still being varied to a certain extent, allowing smooth interpolation and enabling the construction of new samples. To make sure this is satisfied, a variational autoencoder employs the Kullback-Leibler (KL) divergence \cite{kl_divergence}  into its loss function. The KL divergence between two probability distributions measure how much they diverge from each other. Thus, minimizing the KL divergence during training will optimize the probability distribution parameters (mean \(\mu\) and standard deviation \(\sigma\)) to closely resemble that of the target distribution. The KL divergence is mathematically defined as follows:

\begin{equation}
    \sum_{n}^{i=1} \sigma_{i}^{2} + \mu_{i}^{2} - \log(\sigma_{i})-1
\end{equation}

In other words, this loss encourages the encoder to distribute all encodings evenly around the center of the latent space. If the encoder clusters them apart into specific regions away from the origin, it will be penalized. Optimizing the KL divergence loss, combined with the reconstruction loss will allow the generation of a latent space which maintains the similarity of nearby encodings on the local scale via clustering yet is globally densely packed near the latent space origin. In short, this will ensure that the model generates diverse images while maintaining a certain degree of resemblance to the images in the original input. The overall architecture of a variational autoencoder is shown in figure 3.2. 

\begin{figure}[h]
\centering
\includegraphics[width=0.8\textwidth]{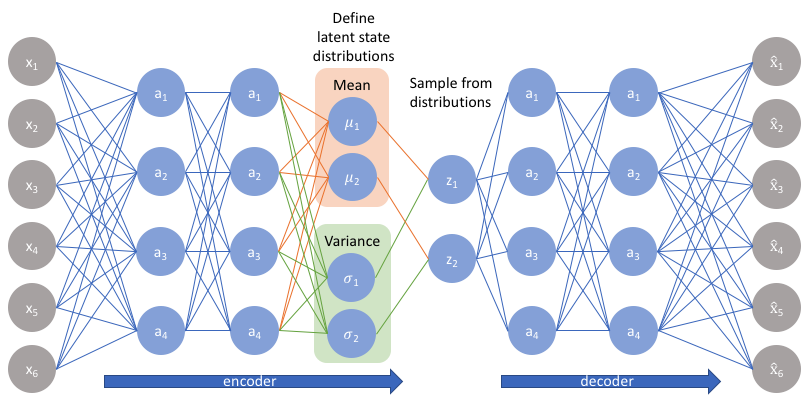}
\caption{A variational autoencoder architecture \cite{jjordan}}
\end{figure}

Since its introduction, there has been an observable amount of uses of variational autoencoders, amongst them are to generate labels and captions to images \cite{vae_captions}, anomaly detection in a variety of of applications \cite{vae_anomaly1}\cite{vae_anomaly2}\cite{vae_anomaly3}, as well as applications in the natural language processing field, such as for semi-supervised text classification \cite{vae_txtcls}, text generation \cite{vae_textgen} and fake news detection \cite{vae_fakenews}.

\subsection{Generative Adversarial Networks (GANs)}

A Generative Adversarial Network - or in short, GAN - is a framework for estimating generative models through an adversarial process where two models are trained simultaneously \cite{ian_gan}: a \textbf{generative model} that captures the data distribution, and a \textbf{discriminative model} that estimates the probability that estimates the probability that a sample came from the training data rather than from a generative model. During training, the generator will try to maximize the probability of the discriminator to make a mistake, i.e. ‘thinking’ that an image comes from the training data instead of the generator, thus ‘fooling’ the discriminator. The training process closely mimics a two-player minimax game. 

Since its introduction, a variety of GAN architectures have been proposed, each being state-of-the-art models for a multitude of applications. Amongst them are Cycle-consistent GAN (CycleGAN) for unpaired image-to-image translation \cite{cyclegan}, deep convolutional GAN (DCGAN) for unsupervised representation learning \cite{dcgan}, and unified GAN (StarGAN) for multi-domain image-to-image translation \cite{stargan}. For our project, we chose to use the StarGAN architecture, primarily because of its ability to perform image-to-image translations for multiple domains using only a single model.

\subsubsection{StarGAN}

A unified model architecture of StarGAN allows simultaneous training of multiple datasets with different domains within a network. Since our project’s main aim is to mitigate biases across three different domains - gender, age, and ethnicity - by generating new images from minority classes, we believe that this architecture is the most appropriate to use. 

\begin{figure}[h]
\centering
\includegraphics[width=0.9\textwidth]{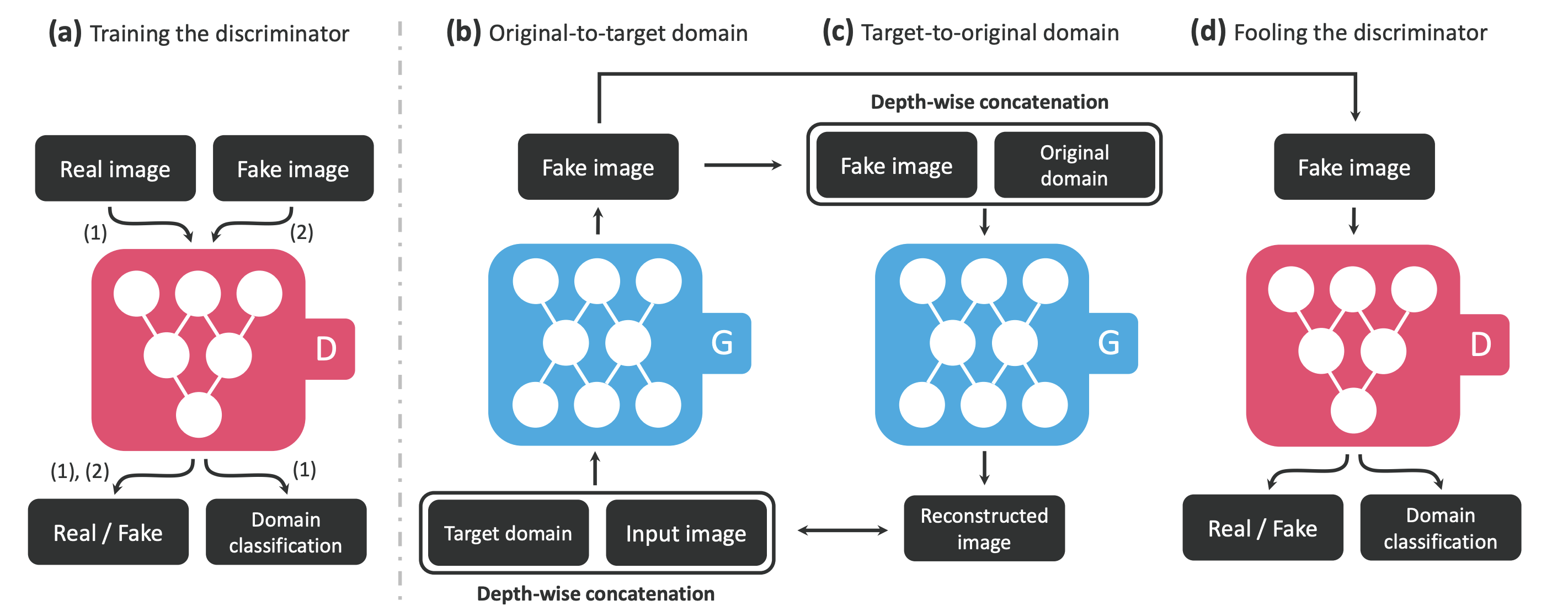}
\caption{Overview of the StarGAN architecture \cite{stargan}}
\end{figure}

The multi-domain translation that is introduced in StarGAN involves using the domain label information as a condition during training. This architecture is novel primarily because its predecessor, CycleGAN, will require you to train \(n(n-1)\) generators if you want to learn all mappings within \(k\) domains. This is highly inefficient and ineffective.

Like all GANs, the StarGAN model consists of two modules: a generator and a discriminator. The difference is that StarGAN’s generator learns mappings among different domains. The discriminator will then try and distinguish between real and fake images and classify real images to its corresponding real domain. During the training phase, the generator \(G\) is trained to translate an input image \(x\) into an output image \(y\) conditioned on the randomly generated target domain label \(c\). This process is formally defined as \(G(x, c) \rightarrow y\).

Simultaneously, an auxiliary classifier is introduced on top of a discriminator \(D\), whose primary function is to classify the real images to its corresponding domain and to classify the fake images to the domain it was conditioned on. As shown in part (a) figure 3.3, the discriminator \(D\) will produce two distributions, \(D : x \rightarrow \left \{  D_{src}(x), D_{cls}(x)\right \}\). So, while \(G\) generates an image \(G(x,c)\), \(D\) will learn to distinguish between real and fake images and produces \(D_{src}(x)\), which is the probability distribution over sources given by \(D\).  On the other hand, \(D_{cls}(x)\) represents the probability distribution over domain labels computed by \(D\).

The StarGAN model employs three main loss functions:
\begin{itemize}
\item
\textbf{Adversarial loss} (\(L_{adv}\)) is a loss function that is present in all GANs. During the training phase, the discriminator \(D\) will try and maximize the error while the generator \(G\) will try to minimize this error, thus simulating the two-player minimax game mentioned previously. This adversarial loss is formally defined as:

\begin{equation}
    L_{adv} = \mathbb{E}_x[\log D_{src}(x)] + \mathbb{E}_{x,c}[\log (1 - D_{src}(G(x,c)))]
\end{equation}

In the above equation, the generator \(G\) is trained to translate an input image \(x\) into an output image \(y\) conditioned on the randomly generated target domain label \(c\), as described previously. This process is demonstrated by (b) in figure 3.3. The discriminator \(D\) will then learn to distinguish between real and fake images and produce the relevant distribution over source data, \(D_{src}(x)\).

\item 

\textbf{Domain classification loss} (\(L_{cls}\)) is associated with classifying and generating images specific to the domains provided from the input labels. The loss function for the domain classification of real images is formally described as follows:

\begin{equation}
    L_{cls}^{r} = \mathbb{E}_{x,c^{\prime}}[-\log D_{cls}(c^{\prime}|x)]
\end{equation}

Here, \(D_{cls}(c^{\prime}|x)\) refers to the probability distribution over domain labels computed by \(D\). By minimizing this, \(D\) will learn to classify a real image \(x\) to its corresponding original domain \(c^{\prime}\). Similarly, \(G\) also tries to minimize the loss function for the domain classification of fake images, denoted by the following equation:

\begin{equation}
    L_{cls}^{f} = \mathbb{E}_{x,c}[-\log D_{cls}(c|G(x,c))]
\end{equation}

Minimizing the above loss function will ensure that \(G\) generates images that can be classified as the target domain \(c\). The optimization of these loss functions are done in (c) in figure 3.3. 

\item

\textbf{Reconstruction loss} (\(L_{rec}\)), also known as the cycle-consistency loss, is used to prevent reconstruction errors after changing specified domains. The reconstruction loss is formally described below: 

\begin{equation}
    L_{rec} = \mathbb{E}_{x,c,c^{\prime}}[\left \| x - G(G(x,c), c^{\prime})) \right \|_{1}]
\end{equation}

This loss function is introduced to guarantee that the translated images preserve the content of its input images while changing only its domain-related parts. While the model reconstructs the original image from the generated image, it calculates the loss between the two. This enforces the model to generate ‘realistic’ images. Formally, generator \(G\) translates input \(x\) to the specified target domain \(c\) and translates it back to the source domain \(c^{\prime}\), as demonstrated in part (d) in figure 3.3. An L1 norm is applied to calculate the loss between the original image \(x\) and the translated image, \(G(G(x,c),c^{\prime}\)).

\end{itemize}

The final loss function for StarGAN’s discriminator (\(L_{D}\)) and generator (\(L_{G}\)) is a combination of losses described above, which is formally denoted as the following:

\begin{equation}
    L_{D} = -L_{adv} + \lambda_{cls}L_{cls}^{r}
\end{equation}

\begin{equation}
    L_{G} = L_{adv} + \lambda_{cls}L_{cls}^{f} + \lambda_{rec}L_{rec}    
\end{equation}

In the equation 3.6 and 3.7, \(\lambda_{cls}\) and \(\lambda_{rec}\) are the model’s hyperparameters whose main objective is to control the relative importance of the domain classification loss and reconstruction loss.


\chapter{Experiments}

This chapter introduces the data preprocessing procedure and some technical details of each data augmentation method including implementation and application. We will also visually display the result of each augmentation technique on the training images in this chapter.  Experiments in this project are performed on a subset of the UTKFace dataset \cite{utkface}, and cross-evaluated on CelebA \cite{celeba} and LFWA+ \cite{lfw}. These datasets will be thoroughly described within this chapter. We will also expound on the preprocessing methods done to the facial recognition datasets and the evaluation procedure that we adopted. 

We pose several research questions (denoted as RQ) that we answer through experiments within this chapter:

\begin{itemize}

\item RQ1: How does each augmentation technique affect the overall classification performance of the model on a test data from the same dataset it was trained on?
\item RQ2: How do the models perform on external facial recognition datasets?
\item RQ3: How does the performance of each model fare against a state-of-the-art attribute classifier?

\end{itemize}

We would expect that a truly unbiased or ‘ideal’ model will have a consistent performance between the different classes in the source dataset as well as on external datasets with the same domain. Through careful experimentation, we hope to obtain a model that performs as closely as the ‘ideal’ model, while providing meaningful analysis and answers to the above questions throughout the experimentation process. The next chapter will contain the results and relevant discussions from the experiments in this chapter.

\section{Data Preprocessing}

For the experiments in this chapter, we have collected a variety of frontal-facing facial images and their respective attributes from three widely-used facial recognition datasets:

\begin{itemize}
    \item 
    The \textbf{UTKFace} dataset \cite{utkface} consists of 20K+ face images in the wild which are readily cropped and aligned, with the respective age, gender, and ethnicity labels. These labels are estimated through the DEX algorithm \cite{dex} and double checked by a human annotator.
    
    \item
    The \textbf{Labeled Faces in the Wild-aligned} (LFWA+) dataset is the preprocessed version of the Labeled Faces in the Wild dataset \cite{lfw} which are aligned by \cite{lfw_aligned}, and contains over 13K face photographs that were designed for studying the problem of unconstrained face recognition, with over 70 attributes including age, gender, and ethnicity. The attributes were externally labeled by Taigman et al. \cite{lfw_labels} through the One-Shot Similarity measure. Positive attribute values indicate the presence of the attribute, while the negative attribute values indicate its absence. The magnitude of the value signifies the degree to which the attribute is present or absent.
    
    \item
    The \textbf{CelebA} dataset \cite{celeba} comes with 200K+ celebrity images with a high diversity across the features. Each image annotated with 40 binary attributes, including age and gender. Unlike the UTKFace and LFWA+ dataset, the CelebA dataset does not come with ethnicity labels. The attributes were annotated using a novel deep learning framework proposed by the authors, which cascades two CNNs, LNet and ANet.
    
\end{itemize}

Below is a summary table on the available annotations within each dataset:
\begin{table}[h]
\centering
\resizebox{0.9\textwidth}{!}{%
\begin{tabular}{|l|l|l|l|l|}
\hline
\multicolumn{1}{|c|}{\multirow{2}{*}{Dataset}}       & \multicolumn{1}{c|}{\multirow{2}{*}{No. of Images}}       & \multicolumn{3}{c|}{Annotations}                \\ \cline{3-5} 
\multicolumn{1}{|c|}{}                      & \multicolumn{1}{c|}{} & Age & Ethnicity & Gender \\ \hline
UTKFace                                     & 13,000+               & $\surd$   & $\surd$         & $\surd$      \\ \hline
Labelled Faces in the Wild, aligned (LFWA+) & 200,000+              & $\surd^{+}$  & $\surd$         & $\surd$      \\ \hline
CelebA                                      & 20,000+               & $\surd^{*}$  &           & $\surd$      \\ \hline
\multicolumn{5}{|l|}{\begin{tabular}[c]{@{}l@{}}* Age labels are binary: young/old\\ + Age labels are categorical: child, youth, middle-aged, senior\end{tabular}} \\ \hline
\end{tabular}%
}
\caption{Summary of the datasets}
\label{tab:my-table}
\end{table}

Given that the UTKFace dataset has the highest number of images and a complete set of annotations, we will choose this dataset as the native dataset to train the model on. We have used the LFWA+ and CelebA dataset for cross-dataset generalization performance evaluation. For each of the images in the dataset, we have performed minimal preprocessing as they are already aligned using dlib's face recognition tool for image alignment \cite{dlib}. To minimize training time and memory consumption, we have cropped them to only contain faces (removed neck and external background) and resized them to 75 x 75 pixels each. Figure 4.1 shows a sample of preprocessed facial images from each dataset.

\begin{figure}[h]
\centering
\includegraphics[width=0.9\textwidth]{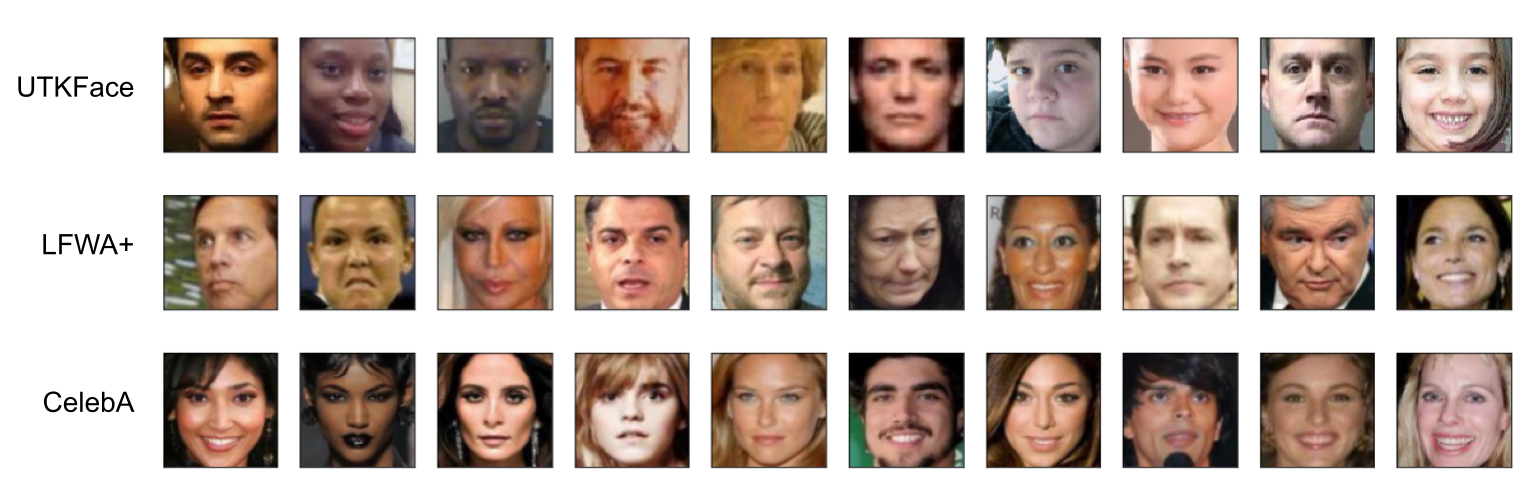}
\caption{Sample images from  UTKFace, LFWA+ and CelebA}
\end{figure}

Furthermore, to understand the degree of bias present, we have performed an initial exploratory data analysis and statistical evaluation on each dataset. The overall class distribution for the UTKFace, CelebA, and LFWA+ datasets can be observed by the graphs on figure 4.2. From the graphs, it is apparent that the attributes in each dataset is prone to class imbalance.

\begin{figure}[h]
\centering
\includegraphics[width=1.0\textwidth]{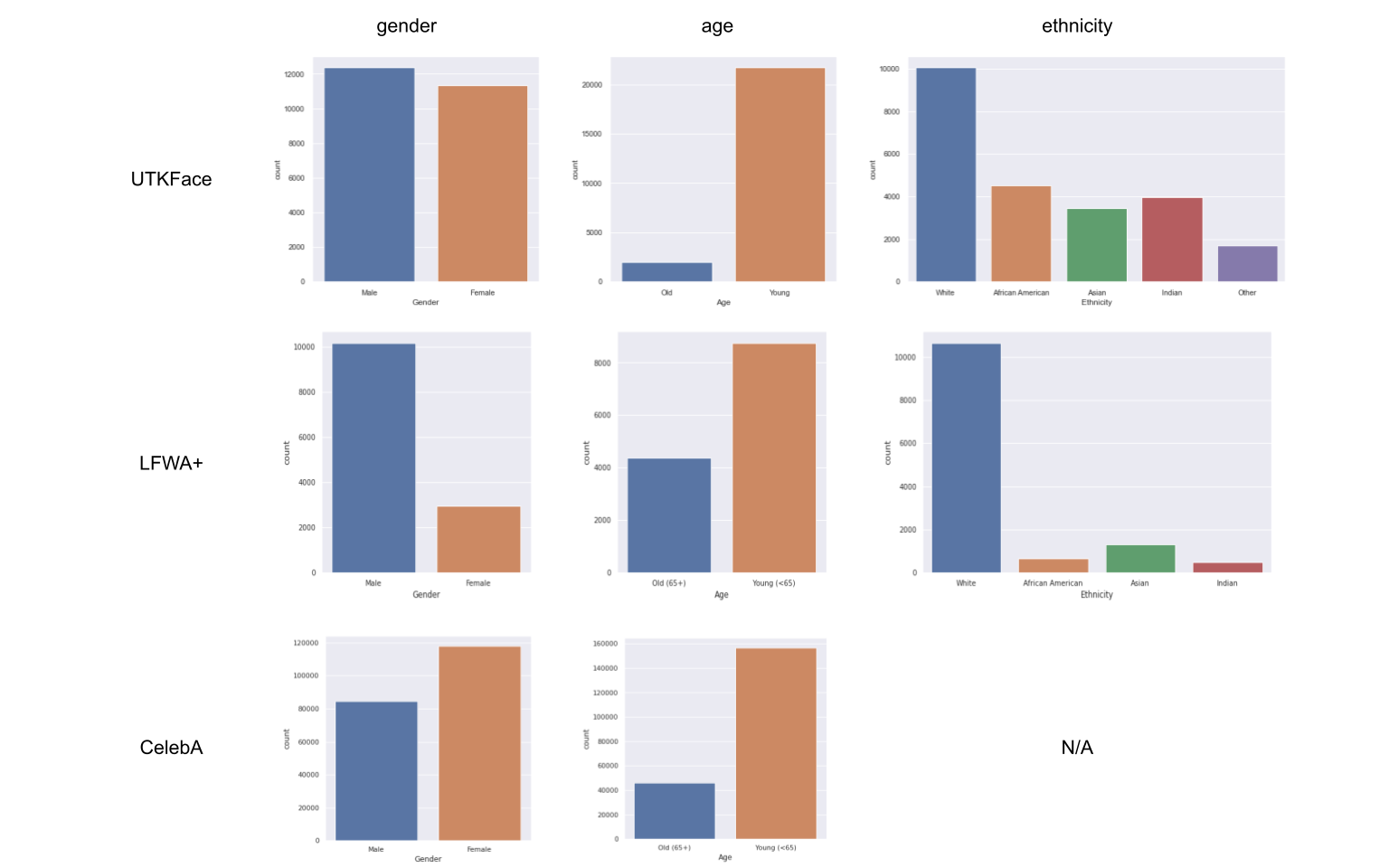}
\caption{Class distributions per attribute for the UTKFace, LFWA+ and CelebA dataset}
\end{figure}

For training the neural network, we used stratified splitting on the UTKFace dataset to obtain the training, validation and test sets, with a respective split of 60/20/20. The justification of the split is that we would like to train the model with as much data as possible while retaining the original dataset variance in the test and validation sets. The splits are reported in table 4.2.

\begin{table}[h]
\centering
\resizebox{0.6\textwidth}{!}{%
\begin{tabular}{|l|l|c|c|c|}
\hline
Attribute                  & Class                 & Train & Validation    & Test \\ \hline
\multirow{2}{*}{Gender}    & Male                  & 7434  & 2456          & 2456 \\ \cline{2-5} 
                           & Female                & 6790  & 2286          & 2286 \\ \hline
\multirow{2}{*}{Age}       & Young (\textless{}65) & 12944 & 4315 & 4315 \\ \cline{2-5} 
                           & Old (65+)             & 1280  & 427           & 427  \\ \hline
\multirow{4}{*}{Ethnicity} & White                 & 6044  & 2015          & 2015 \\ \cline{2-5} 
                           & Black                 & 2718  & 906           & 906  \\ \cline{2-5} 
                           & Asian                 & 2060  & 687           & 687  \\ \cline{2-5} 
                           & Indian                & 2385  & 795           & 795  \\ \hline
\end{tabular}%
}
\caption{UTKFace dataset train/validation/test split statistics}
\label{tab:my-table}
\end{table}

Since the labels of each dataset do not agree with each other, we decided to preprocess the attributes further. Unlike the binary and categorical labels in the CelebA and UTKFace dataset, the LFWA+ dataset assigns a positive or negative numerical value representing how present the attribute is in each image. Therefore, we took the attribute with the highest positive value in each class and assigned a categorical value label to that attribute. For example, the highest positive ethnicity label value is ‘white’, we will assign the ethnicity attribute of that image to be 0, which is the respective categorical label for that specific ethnicity. After preprocessing, the attributes within the LFWA+ dataset adopt a categorical labeling system.

Additionally, CelebA has adopted a binary labeling of ‘old’ and ‘young’, LFWA+ has adopted a categorical labeling of ‘child’, ‘youth’, ‘middle-aged’, and ‘senior’, while UTKFace has the exact numerical age value. For the purpose of this project, keeping in mind efficiency and feasibility of the implementation, we decided to follow CelebA’s binary labeling. Additionally, having two classes that are very diverse will allow a more noticeable image-to-image translation by the StarGAN. Therefore, we have assigned the label ‘old’ for anyone over the age of 65 in UTKFace and for anyone with the ‘senior’ label in LFWA+. We have labeled the remaining population as ‘young’. 

Furthermore, there were five ethnicities available in the UTKFace dataset - white, black, asian, indian, and other - however, we decided to remove the ‘other’ ethnicity group mainly because it contains images of people with assorted ethnicities, thus they do not share as many invariant features with each other as the other ethnicity groups do. This also helps the ethnicity labels in the UTKFace dataset to agree with the ones on the LFWA+ dataset, which do not have the ‘other’ group. In addition to that, the removal of the ‘other’ ethnicity group is also done to reduce the training time. After all the image and attribute preprocessing have been implemented, we calculated an average face from all of the images within each class by taking the mean of the vectorized images, visualized in figure 4.3.

\begin{figure}[h]
\centering
\includegraphics[width=1.0\textwidth]{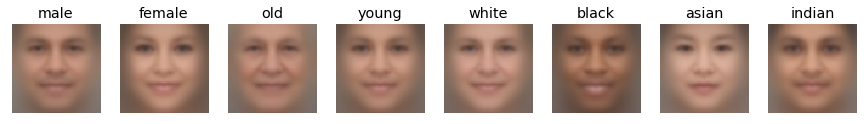}
\caption{Visualization of the mean face vector for each class}
\end{figure}

For external testing, we have randomly sampled a class-wise balanced set of preprocessed images from the CelebA and LFWA+ datasets. This is done such that any discrepancies in the evaluation metrics can be solely attributed to the model performance, and not caused by other factors primarily the class imbalances within the dataset.  Table 4.3 outlines the number of test images used in each class from each dataset.

\begin{table}[h]
\centering
\resizebox{0.5\textwidth}{!}{%
\begin{tabular}{|l|l|c|c|}
\hline
Attribute                  & Class                 & LFWA+ & \multicolumn{1}{l|}{CelebA} \\ \hline
\multirow{2}{*}{Gender}    & Male                  & 2000  & 4000                        \\ \cline{2-4} 
                           & Female                & 2000  & 4000                        \\ \hline
\multirow{2}{*}{Age}       & Young (\textless{}65) & 4000  & 4000                        \\ \cline{2-4} 
                           & Old (65+)             & 4000  & 4000                        \\ \hline
\multirow{4}{*}{Ethnicity} & White                 & 800   & -                           \\ \cline{2-4} 
                           & Black                 & 800   & -                           \\ \cline{2-4} 
                           & Asian                 & 800   & -                           \\ \cline{2-4} 
                           & Indian                & 800   & -                           \\ \hline
\end{tabular}%
}
\caption{Number of test images per class from each dataset}
\label{tab:my-table}
\end{table}

\section{Data Augmentation}

In this section, we will describe the implementation details of the different data augmentation techniques described in chapter 3. All of the augmentation processes in this section are done on a single 10 GB NVIDIA Tesla K40m GPU in a virtual environment under Scientific Linux version 7.8 (Nitrogen). All data augmentation methods are solely performed on the training set, while the validation and test sets are kept constant.

To perform undersampling, we have picked the classes with the lowest number of training images in each attribute to obtain a balanced training set. Thus, each class within the same attribute would have an equal number of instances to the class with the least number of instances. After undersampling, the gender, age, and ethnicity attribute has 6790, 1280, and 2060 instances per class respectively. 

For geometric transformations, we have used a rotation with a maximum rotation degree of 10, zooming with a factor between 1.1 and 1.2, and horizontal flipping. These values were chosen to preserve the natural properties of the facial recognition dataset, namely maintaining a roughly vertical facial alignment, not cropping out essential features such as eyebrows and bottom lip through excessive zooming, etc. The effects of applying geometric transformation on an image from the UTKFace dataset is shown in figure 4.4.

\begin{figure}[h]
\centering
\includegraphics[width=0.8\textwidth]{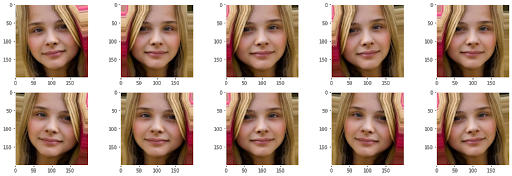}
\caption{Geometrically transformed image from the UTKFace dataset}
\end{figure}

For the next augmentation technique, we have implemented a simple variational autoencoder using PyTorch \cite{pytorch} resembling the architecture shown in figure 4.5. The encoder and decoder model contains a single fully-connected layer each. The encoder network turns the input samples into two parameters in a latent space: the vector of means \(\mu\) and the vector of standard deviations \(\sigma\). In the sampling layer, we will use these vectors and a random normal tensor \(\varepsilon\) to randomly sample similar points \(z\), mathematically denoted as the following equation:

\begin{equation}
    z = \mu + e^\sigma \times \varepsilon 
\end{equation}

Then, we built a decoder network that maps these randomly sampled latent space points back to the original input data.

\begin{figure}[h]
\centering
\includegraphics[width=0.8\textwidth]{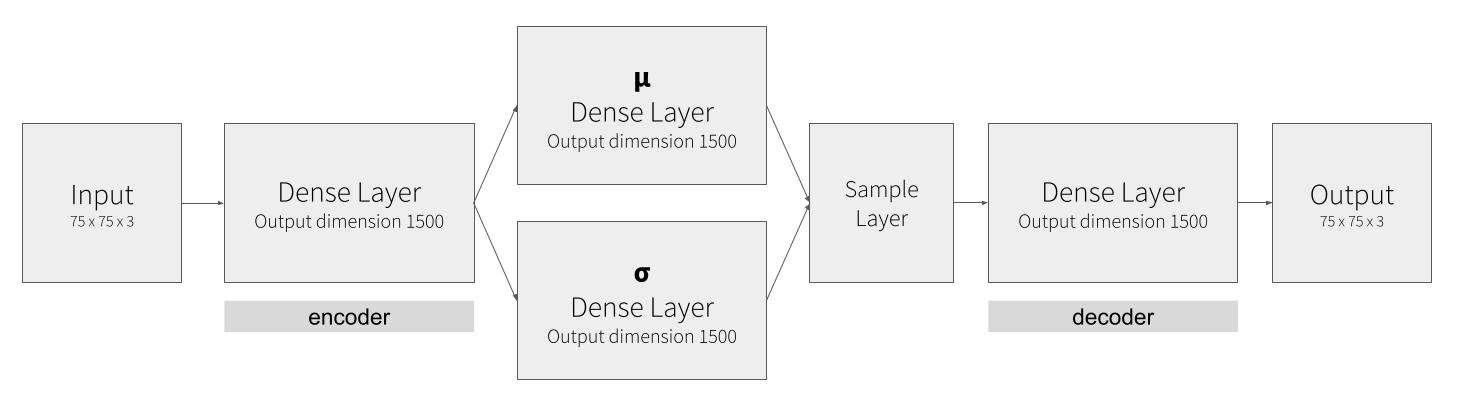}
\caption{The variational autoencoder architecture used in this project}
\end{figure}

The parameters of the variational autoencoder model are trained via two loss functions: the reconstruction loss and the KL divergence \cite{kl_divergence}. The reconstruction loss is the mean squared error between the output and input, and it ensures that the decoded samples match the initial inputs. The KL divergence between the learned latent distribution and the prior distribution acts as a regularization term. Optimizing the KL divergence helps learning well-formed continuous latent spaces as well as reducing overfitting to the training data.

Finally, we were able to train our variational autoencoder model on the preprocessed facial images in the UTKFace dataset for over 20 epochs with a batch size of 128. We have used the Adam optimizer \cite{adam} with a learning rate of 0.001. The primary reason for this  configuration is to optimize accuracy while also keeping training time at a minimum. We computed the latent vector for each class in each attribute within the UTKFace dataset and mapped an image onto each latent vector. The generated images obtained through the use of our variational autoencoder across multiple domains is shown in figure 4.6.

\begin{figure}[h]
\centering
\includegraphics[width=0.8\textwidth]{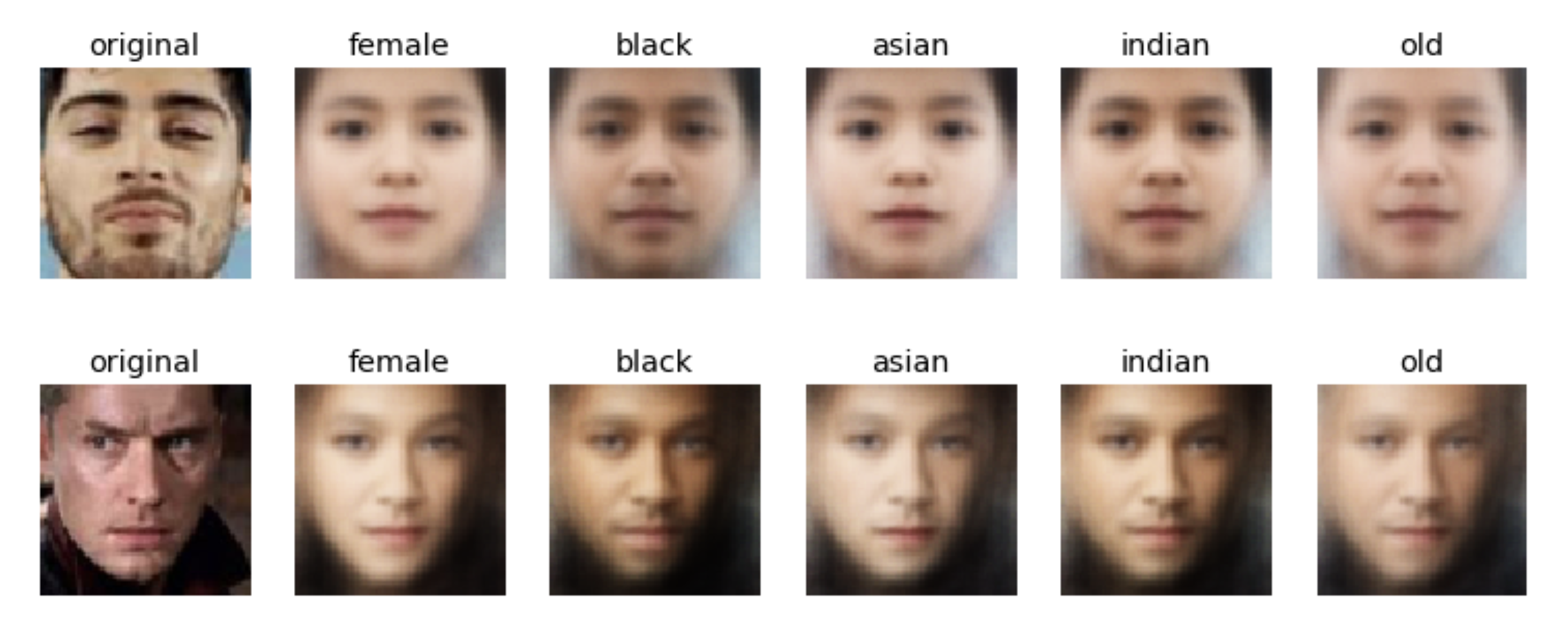}
\caption{Images generated by the variational autoencoder}
\end{figure}

The final augmentation technique to be evaluated in this project is image generation through a Generative Adversarial Network (GAN). To be able to obtain a GAN that generates good quality images, a large amount of data is required. For the ‘old’ class in the ‘age’ attribute, we have less than 2000 training data. Therefore, we used the previous geometric transformation method to generate the remaining 720 images. As mentioned previously, we have chosen StarGAN because of its ability to perform image-to-image translations for multiple domains by training only a single model. This model allows simultaneous training of multiple datasets with different domains within a network. We made use of the implementation provided by the authors of the original paper \cite{stargan_imp}.

We train our StarGAN on all three attributes with a total of eight different classes: male, female, young, old, white, black, asian, and indian. We have used 2000 images for each class during training. Due to hardware limitations, we only managed to train the network for 20,000 iterations with a batch size of 16. We have kept the images to a size of 75 x 75 pixels. To be able to visually monitor the performance of the model during training, we have set the model to save a checkpoint after every 1000 iterations and display a sample of translated images from a single reference image to its respective domains. The training process took about 50 hours for the specified parameter settings. The final model is able to generate realistic images from a single source image across eight different domains through latent-guided synthesis, as shown in figure 4.7. Through quick visual examination, we can observe that the images generated by the StarGAN are of high perceptual quality and resemble realistic human faces.

\begin{figure}[h]
\centering
\includegraphics[width=0.9\textwidth]{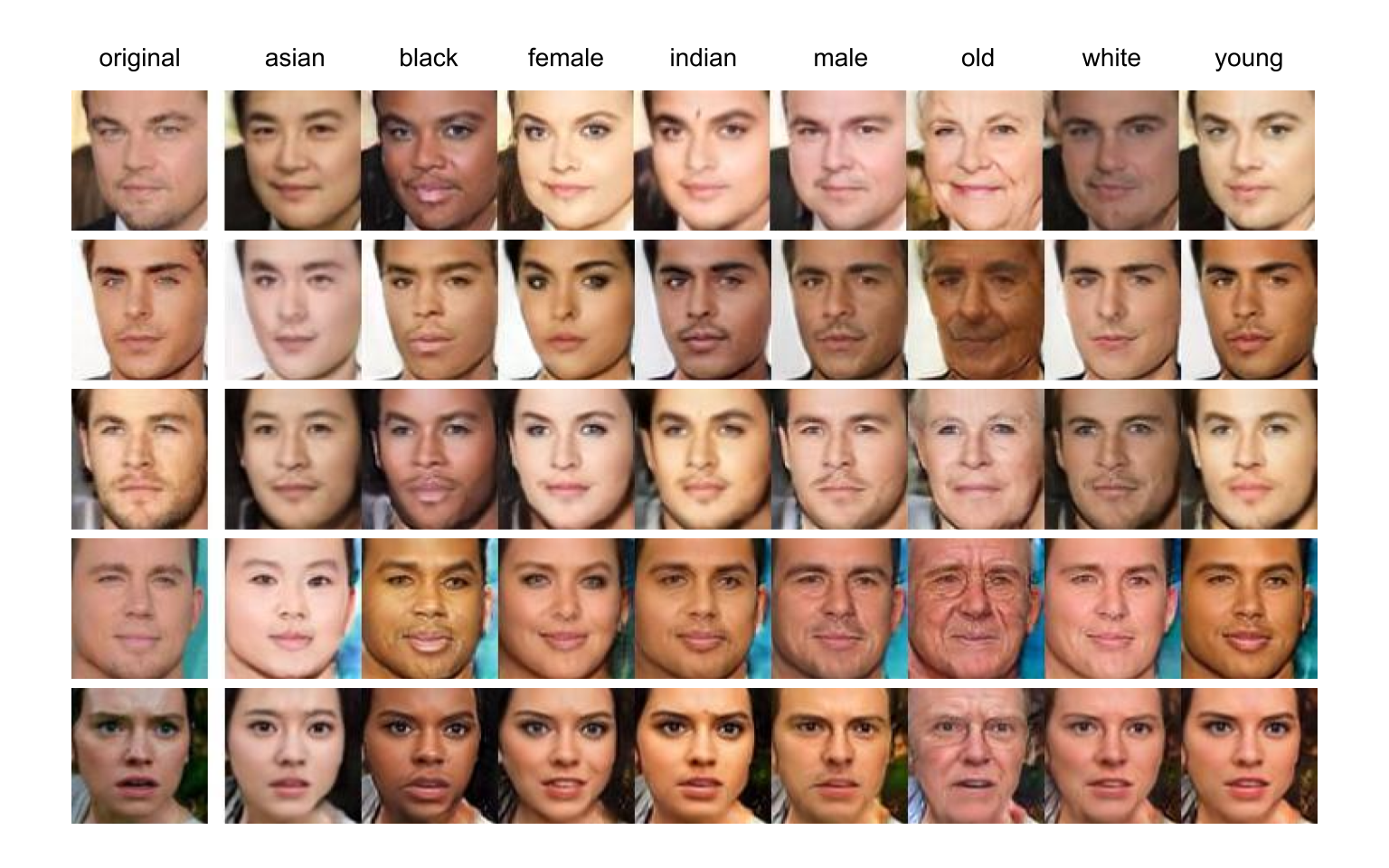}
\caption{Images generated by the StarGAN through latent-guided synthesis}
\end{figure}

It is also important to note that due to the random nature of the image generation, diversifying a particular domain will also unintentionally lead to diversifying multiple domains. For example, generating a ‘female’ image will generate an image that belongs to a different ‘ethnicity’ group than the original source image. This helps prevent homogenization of a particular class and the loss of in-class variance.

Contrary to undersampling, every class in each attribute will have the same number of instances as the majority class after generating new training images through geometric transformations, the variational autencoder and the StarGAN. Each class within the gender, age, and ethnicity attribute will now have 7434, 12944, and 6044 instances respectively.

\section{Experimental Setup}

\subsection{Network Architecture}

For each of the dataset obtained as a result of the aforementioned augmentation techniques, we have built an InceptionV3 model \cite{inceptionv3} using TensorFlow \cite{tensorflow} to train them on. Choosing a state-of-the-art pretrained model is favored above building our own neural network so that any performance discrepancies within the result can solely be attributed by the dataset being biased, and not the model’s capability of classification. 

Furthermore, Zebin Jiang \cite{zebin} has compared three state-of-the-art network architectures for gender classification, namely VGG16 \cite{vgg16}, InceptionV3 \cite{inceptionv3} and ResNet50 \cite{resnet50}. In the paper, it was found that VGG16 performs gender classification the best, with an accuracy of 95\%. However, training a VGG16 takes around 37 seconds per epoch. Therefore, to reduce the training time, we decided to use InceptionV3, which takes 2 seconds per epoch to train and has an accuracy of 91\%. This choice is made to make a compromise between accuracy and efficiency.

The structure of the InceptionV3 model is shown below:

\begin{figure}[h]
\centering
\includegraphics[width=0.9\textwidth]{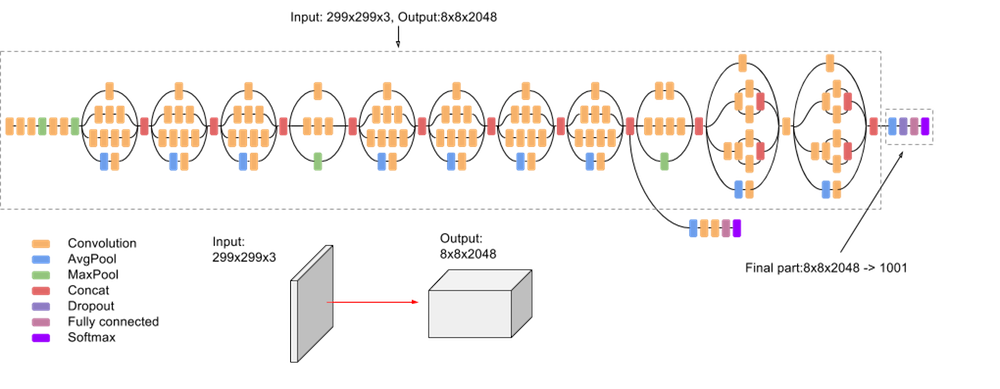}
\caption{InceptionV3 architecture \cite{iv3_arch}}
\end{figure}

In short, InceptionV3 is a state-of-the-art 42-layer deep convolutional neural network architecture from the Inception family that makes several improvements such as using Label Smoothing, Factorized 7 x 7 convolutions, and the use of an auxiliary classifier to propagate label information lower down the network. It also uses batch normalization for layers in the sidehead. 

For this project, we have used an InceptionV3 network that was pre-trained on the ImageNet dataset \cite{imagenet}. To train the network as an attribute classifier on our dataset, we have replaced the top layers by the following trainable layers:
\begin{enumerate}
\itemsep-0.5em
    \item 2D Global Average Pooling 
    \item Fully connected layer with output dimension 1024 and ReLU activation
    \item Dropout with probability 0.5
    \item Fully connected layer with output dimension 512 and ReLU activation
    \item Fully connected layer with output dimension \(x\) and softmax activation
\end{enumerate}

The \(x\) above signifies the number of classes within each attribute, which is 2 for gender (male, female), 2 for age (young, old), and 4 for ethnicity (white, black, asian, indian). These top layers will be trained on the augmented images.

We have trained each model on the preprocessed augmented images with a batch size of 64 for 25 epochs to avoid overfitting, while monitoring the validation loss (categorical cross-entropy) and accuracy at every iteration. We have used the Stochastic Gradient Descent optimizer with a learning rate of 0.0001 and a momentum of 0.9. These parameters were chosen to maximize accuracy however still maintain a relatively efficient training process.

\subsection{Performance Evaluation}

It is critical to evaluate the performance of the models trained on the augmented versions of the UTKFace dataset using metrics such as F1-score, as classification accuracy is known to fail on classification problems with a skewed class distribution. This is because classification accuracy is initially designed by practitioners on datasets with an equal class distribution. Nevertheless, we will report the per-class accuracy within each attribute for the ease of comparison with the state-of-the-art attribute classification model. We aim to investigate to what extent each augmentation method improves the model’s performance on minority classes and on external datasets. To do this, we ran the classifier on a test set from the original UTKFace dataset and used the resulting performance as the baseline comparison. The complete results of the experiment will be reported and discussed in the next chapter.

\subsubsection{Evaluation on Native Dataset}

To answer RQ1, we would evaluate the performance of each model on the test set from the native dataset, i.e. the dataset that the model was trained on, which was UTKFace. As highlighted in the previous section, we moved 20\% of the images from the original unaugmented dataset for testing. The detailed statistics of the split can be found in table 4.2. We then ran the classifiers on this test set. From the results of this experiment, we hope to see whether or not the model performs uniformly on the majority and minority classes. This will be shown by any discrepancies in per-class accuracies and F1-scores within a particular attribute.

\subsubsection{Cross-Dataset Generalization}

Similarly, we evaluated the performance of the classifiers on other facial recognition datasets that were not used for training, CelebA and LFWA+, to answer RQ2. A truly balanced and unbiased model should be able to generalize on external datasets of the same domain. In figure 4.3 in the previous section, we have calculated and visualized the average faces in the UTKFace dataset. We have also calculated the average faces for the CelebA and LFWA+ datasets and calculated the difference from each average face vector to the UTKFace average face vector. This is done by calculating the mean squared error and the structural similarity index (SSIM) \cite{ssim} between the vectors. The mean squared error measures the difference between each pixel within the images, while the structural similarity index attempts to model the perceived change in the structural information of the image. If two images are identical, the mean squared error between them will be 0 and the structural similarity index will be 1. From this, we aim to be able to investigate the correlation between the similarity of the average images between the native and external test dataset has an effect on the generalizability of a model to the test dataset. 

Finally, from the results of the cross-dataset performance evaluation, we will be able to gauge a model’s ability to generalize. A high performance on the native dataset and a low performance on the external dataset may signify that the model has learned the intrinsic biases within the training dataset. We will also be able to see whether or not the model is biased towards the majority class by examining the per-class accuracies and F1-scores.

\subsubsection{Comparison with State-Of-The-Art}

Lastly, we will compare the best model for each attribute on a state-of-the-art attribute classifier trained on the FairFace dataset \cite{fairface} discussed in chapter 2. The results of this evaluation will help us answer RQ3. We want to be able to investigate how a model trained on our augmented datasets will fare against the current state-of-the-art balanced dataset.  The authors of the FairFace dataset have trained a simple classifier based on ResNet-34 and have successfully shown that the model trained from the FairFace dataset is significantly more accurate on various face recognition datasets and the accuracy is consistent between race and gender groups. 

For a fair comparison with our models and to obtain meaningful results, we have obtained the model trained on FairFace that was published by the original authors and ran the classifier on the same test set we have used for our native and cross-dataset evaluation. We will report the similar metrics as the previous experiments in order to gauge the consistency and generalizability of the FairFace model on our test set. We also made sure not to change any image or attribute preprocessing method. 


\chapter{Results}

In the previous chapter, we have thoroughly described the technical implementation details of the various data augmentation techniques as well as the evaluation methods we have chosen. In this chapter, we will report in detail and analyze the results of the experiments. We will also aim to answer the questions defined in the previous chapter.

\section{Evaluation and Discussion}

\subsection{RQ1: Performance Evaluation on Native Dataset}

The detailed results of performing classification on the UTKFace test set on each of the different models are shown in table 5.1. The highest average accuracy and F1-score for each class is shown in bold. From a quick evaluation of the table, we can notice that almost all of the augmentation techniques performed on the UTKFace training set increase the overall performance of the model. Furthermore, we can also notice that the best models trained on augmented data also leads to a more consistent performance, with a standard deviation of no more than 0.02 between the accuracies and F1-scores of the classes within each attribute. This is a significant drop from the baseline that has an average standard deviation of almost 0.1. This shows that the augmentation techniques more or less alleviates the bias problem in the baseline model. 

\begin{table}[h]
\centering
\resizebox{\textwidth}{!}{%
\begin{tabular}{|l|l|c|c|c|c|c|c|c|c|c|c|}
\hline
\multirow{2}{*}{Attribute} &
  \multirow{2}{*}{Class} &
  \multicolumn{2}{c|}{Baseline} &
  \multicolumn{2}{c|}{Undersampling} &
  \multicolumn{2}{c|}{Geometric} &
  \multicolumn{2}{c|}{Var. Autoencoder} &
  \multicolumn{2}{c|}{StarGAN} \\ \cline{3-12} 
                           &        & Acc.  & F1 & Acc.  & F1 & Acc.           & F1       & Acc.  & F1 & Acc.           & F1       \\ \hline 
\multirow{2}{*}{Gender}    & Male   & 0.910 & 0.900    & 0.916 & 0.900    & 0.882          & 0.880          & 0.932 & 0.880    & \textbf{0.944} & \textbf{0.910} \\ \cline{2-12} 
                           & Female & 0.870 & 0.900    & 0.888 & 0.900    & 0.873          & 0.880          & 0.825 & 0.870    & \textbf{0.891} & \textbf{0.900} \\ \hline 
\multirow{2}{*}{Age}       & Old    & 0.604 & 0.740    & 0.913 & 0.850    & \textbf{0.936} & \textbf{0.920} & 0.555 & 0.710    & 0.801          & 0.880          \\ \cline{2-12} 
                           & Young  & 0.981 & 0.830    & 0.766 & 0.830    & \textbf{0.890} & \textbf{0.910} & 0.983 & 0.810    & 0.975          & 0.900          \\ \hline 
\multirow{4}{*}{Ethnicity} & White  & 0.838 & 0.810    & 0.634 & 0.650    & 0.846          & 0.820          & 0.892 & 0.770    & \textbf{0.854} & \textbf{0.840} \\ \cline{2-12} 
                           & Black  & 0.862 & 0.870    & 0.818 & 0.810    & 0.894          & 0.900          & 0.862 & 0.790    & \textbf{0.886} & \textbf{0.900} \\ \cline{2-12} 
                           & Asian  & 0.752 & 0.800    & 0.746 & 0.760    & 0.824          & 0.880          & 0.670 & 0.730    & \textbf{0.892} & \textbf{0.890} \\ \cline{2-12} 
                           & Indian & 0.800 & 0.770    & 0.708 & 0.690    & 0.846          & 0.830          & 0.462 & 0.560    & \textbf{0.854} & \textbf{0.860} \\ \hline 
\multicolumn{12}{|l|}{* the models with the best performance (average accuracy and F1-score) are highlighted in bold}                                  \\ \hline
\end{tabular}%
}
\caption{Classification performance statistics of each model on the UTKFace test set}
\label{tab:my-table}
\end{table}

However, this trend is not observed on the models trained on the dataset augmented through variational autoencoders. In fact, all these models have a relatively poor performance, yielding lower accuracies than the baseline and increasing the disparity between the performance of the majority class and the minority class. This is probably because the images generated by variational autoencoders are less realistic and far more blurry with softer edges, thus diminishing the appearance of key features. To highlight this observation, we have presented a comparison between the images generated by the variational autoencoder and the images generated by the StarGAN in figure 5.1.

\begin{figure} [h]
    \centering
    \includegraphics[width=0.8\textwidth]{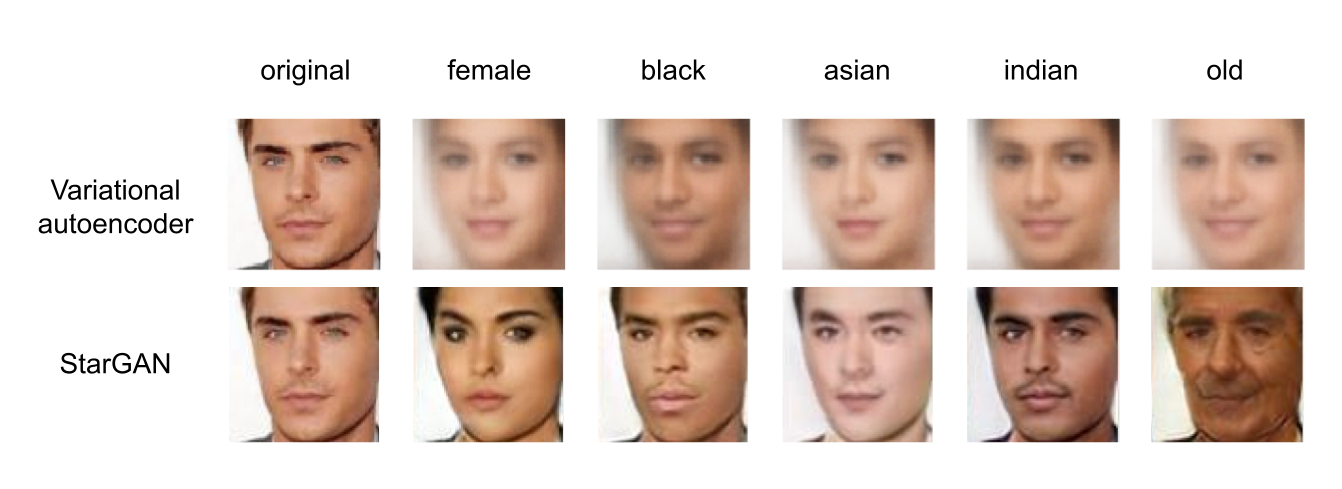}
    \caption{Comparison between the images generated by the variational autoencoder and StarGAN}
    \label{fig:my_label}
\end{figure}

Another common trend that can be observed is that the models trained on undersampled data notably reduces the performance of the majority class. This can be attributed to the fact that undersampling based on the least class will discard a significant portion of the majority class, removing the in-class variance and potentially throwing away potentially useful information.

For the gender attribute, we can see that using trained StarGAN to generate additional training images yields the highest performance with respect to accuracy and F1-score, with an average accuracy of 91.75\%. However, training on the geometrically transformed dataset leads to a more uniform performance across the classes within the gender attribute, with an accuracy standard deviation of 0.006 comapred to the StarGAN model's 0.038. Regardless, for the majority of the augmentation techniques, the F1-scores for the classification of the gender attribute show a consistent performance across the classes, even the baseline model. From this, we can deduce that the class imbalance in the gender attribute is not severe enough to cause the model to be biased.

Furthermore, we have noticed that training our model on the geometrically transformed training set produces the best classification performance on the age attribute, with an accuracy of 91.30\%. This model also notably improved the performance consistency between the classes, dropping the accuracy standard deviation from 0.189 in the baseline to 0.023. Training on the StarGAN-generated images does increase the classification performance of the minority class, however it still lags far behind the majority class. Additionally, we get to observe how undersampling significantly reduces the performance of the majority class particularly in this attribute, from 98.1\% in the baseline to 76.6\%. If we recall from the statistical evaluation of the UTKFace dataset, there is a huge difference in the number of instances in the majority class and the minority class. Removing thousands of instances in the ‘young’ class in undersampling will reduce the variance in the class, and variance is particularly important in this class because a ‘young’ person covers anyone from birth to someone who is middle-aged.

Amongst the augmented variants of the UTKFace dataset, training an ethnicity classifier on the StarGAN-generated images yield the best overall performance. This model has a classification accuracy of 87.2\% and shows a consistent performance across the different classes within the attribute, with an accuracy standard deviation of only 0.017. Furthermore, the model also increases the accuracies for each class from the baseline. This trend can also be observed for the model trained on the geometrically transformed dataset. Similarly to the age attribute, undersampling reduces the accuracy of the majority class by a notable amount, from around 83.6\% to 63.4\%. There is also a sizable class imbalance in this attribute, with the ‘white’ class containing around 6000 instances while the remaining classes contain only about 2000 instances each. Thus, removing over half of the instances in the ‘white’ class in training deteriorates the model’s ability to distinguish ‘white’ faces within the test set.

Overall, we can conclude that training an attribute classifier on a balanced training set through augmentation more or less improves the overall performance of the model and also leads to a more consistent performance across the different classes within the attribute. For most of the attributes, StarGAN proves to be the most effective in improving the performance as well as mitigating the performance discrepancies between the majority and minority classes. However, it is also important to note that the model trained on the geometrically transformed images performs very closely. In fact, it produced the best performing classifier on the age attribute. Considering that training a StarGAN and generating the images take a total of almost 50 hours while geometrically transforming images take only a few minutes, augmenting the dataset using geometric transformations can be the best compromise between accuracy and efficiency.

\subsection{RQ2: Cross-Dataset Generalization}

In the previous chapter, we have mentioned briefly about measuring the similarity between the average face vector of the UTKFace dataset and the LFWA+ and CelebA dataset. This was done by calculating the mean squared error and structural similarity index between the vectors. If two datasets contain similar images, they will have a mean squared error close to 0 and a structural similarity index close to 1. The values are reported in table 5.2. 

\begin{table}[h]
\centering
\resizebox{0.7\textwidth}{!}{%
\begin{tabular}{|l|c|c|}
\hline
       & Mean Squared Error & Structural Similarity Index (SSIM) \\ \hline
LFWA+  & 0.0108             & 0.8050                             \\ \hline
CelebA & 0.0258             & 0.7172                             \\ \hline
\end{tabular}%
}
\caption{Similarity comparison of the average face vector of LFWA+ and CelebA to the average face vector of UTKFace}
\label{tab:my-table}
\end{table}

From table 5.2, we can observe that the average face vector of the training dataset, UTKFace, is more similar to the LFWA+ dataset than CelebA. However, the differences in the mean squared error and structural similarity index is not too huge, signifying that the model has the potential to generalize over both datasets.

\subsubsection{Performance on the LFWA+ Dataset}

\begin{table}[h]
\centering
\resizebox{\textwidth}{!}{%
\begin{tabular}{|l|l|c|c|c|c|c|c|c|c|c|c|}
\hline
\multirow{2}{*}{Attribute} &
  \multirow{2}{*}{Class} &
  \multicolumn{2}{c|}{Baseline} &
  \multicolumn{2}{c|}{Undersampling} &
  \multicolumn{2}{c|}{Geometric} &
  \multicolumn{2}{c|}{Var. Autoencoder} &
  \multicolumn{2}{c|}{StarGAN} \\ \cline{3-12} 
                           &        & Acc.  & F1    & Acc.  & F1    & Acc.           & F1             & Acc.  & F1    & Acc.           & F1             \\ \hline
\multirow{2}{*}{Gender}    & Male   & 0.887 & 0.850 & 0.882 & 0.840 & 0.866          & 0.880          & 0.896 & 0.820 & \textbf{0.902} & \textbf{0.900} \\ \cline{2-12} 
                           & Female & 0.808 & 0.840 & 0.776 & 0.820 & 0.888          & 0.880          & 0.708 & 0.780 & \textbf{0.917} & \textbf{0.900} \\ \hline
\multirow{2}{*}{Age}       & Old    & 0.400 & 0.330 & 0.710 & 0.760 & \textbf{0.753} & \textbf{0.810} & 0.158 & 0.270 & 0.577          & 0.710          \\ \cline{2-12} 
                           & Young  & 0.990 & 0.710 & 0.848 & 0.790 & \textbf{0.890} & \textbf{0.830} & 0.977 & 0.690 & 0.942          & 0.800          \\ \hline
\multirow{4}{*}{Ethnicity} & White  & 0.936 & 0.640 & 0.800 & 0.640 & 0.984          & 0.680          & 0.892 & 0.610 & \textbf{0.950} & \textbf{0.700} \\ \cline{2-12} 
                           & Black  & 0.776 & 0.780 & 0.788 & 0.740 & 0.624          & 0.730          & 0.784 & 0.680 & \textbf{0.790} & \textbf{0.810} \\ \cline{2-12} 
                           & Asian  & 0.384 & 0.530 & 0.888 & 0.610 & 0.392          & 0.550          & 0.208 & 0.320 & \textbf{0.650} & \textbf{0.700} \\ \cline{2-12} 
                           & Indian & 0.356 & 0.440 & 0.424 & 0.480 & 0.544          & 0.550          & 0.172 & 0.240 & \textbf{0.575} & \textbf{0.540} \\ \hline
\multicolumn{12}{|l|}{* the models with the best performance (average accuracy and F1-score) are highlighted in bold}                                    \\ \hline
\end{tabular}%
}
\caption{Classification performance statistics of each model on the LFWA+ test set}
\label{tab:my-table}
\end{table}

Based on the structural similarity between the average face vector of the UTKFace and LFWA+, we would expect that the performance of our models on the LFWA+ dataset will be quite similar to the performance on the UTKFace dataset. By looking at the detailed model classification performance statistics in table 5.3, we can observe that the best models for each attribute in the LFWA+ dataset correspond to the ones in the UTKFace dataset. However, the performance is slightly worse overall and less balanced, especially for the age and ethnicity attribute. This is as expected, because the models have not seen a single instance from this dataset. 

Nevertheless, the overall performance and consistency between the classes increase from the baseline as we apply most of the augmentation techniques on the training set. This shows that the models that successfully alleviate biases in their source dataset also generalizes better. Similar to the results of the classifier obtained on the UTKFace dataset, we observe a trend where undersampling reduces the performance of the majority class, however the effect is not as severe. Training on images generated by variational autoencoders also leads to the worst performance, reducing the accuracy of each class within the attributes and increasing the gap between the performance of the majority and minority classes.

For the gender attribute, the best performing model is obtained through training the model on the StarGAN-generated images, with an accuracy of 91\%. The performance of the model on the LFWA+ dataset is not too far behind when compared to the performance on the UTKFace dataset, which had an accuracy of 91.7\%. In addition to that, a standard deviation of less than 0.1 between the class accuracies and F1-scores shows that the performance is consistent among the classes within the attribute. Training the model on the geometrically transformed images also yields a similar performance, differing by only a few points at 87.7\%. 

Just like the UTKFace dataset, the best performing model for the age attribute is obtained through training the model on geometrically transformed images, with an accuracy of 82.2\%. This is quite a significant drop from the accuracy on the UTKFace dataset, which was 91.30\%. Additionally, although the model obtained through this method reduces the performance of the majority class from the baseline, it increases the accuracy of the minority class by almost two-fold, from around 40\% to 75\%. The standard deviation of the accuracies also reduced from the baseline, from 0.42 to less than 0.1. Furthermore, the model obtained by training the on undersampled data and on StarGAN-generated images also improves the performance of the minority class from the baseline, however there are still some noticeable discrepancies between the performances of each class. 

Finally, the model that has the highest classification performance on the ethnicity attribute is the model that was trained on images generated by the StarGAN, with an overall accuracy of 74.1\%. Out of all the attributes, this is the highest drop in performance when compared to the model’s performance on the UTKFace dataset, which was 87.2\%. We also noticed that the performance consistency between the classes are arguably poor, showing favor towards the majority class. This is statistically confirmed by a standard deviation of the accuracies of 0.143 in the best model. Regardless, this is a significant improvement from the baseline standard deviation of 0.288. Additionally, training on undersampled data and geometrically transformed images increase the overall performance and consistency from the baseline, however the performance on the minority classes are still very poor, with accuracies of only 40-50\%. 

Thus, we can deduce that training the model on an augmented version of the source dataset yields a better overall performance and generalization capability on the LFWA+ dataset. In addition to that, it also leads to a model that is less biased towards the majority class. Therefore, we can conclude that training the model on a balanced dataset obtained through augmentation mitigates the effect of intrinsic biases within the source dataset to a notable extent.

\subsubsection{Performance on the CelebA Dataset}

\begin{table}[h]
\centering
\resizebox{\textwidth}{!}{%
\begin{tabular}{|l|l|c|c|c|c|c|c|c|c|c|c|}
\hline
\multirow{2}{*}{Attribute} &
  \multirow{2}{*}{Class} &
  \multicolumn{2}{c|}{Baseline} &
  \multicolumn{2}{c|}{Undersampling} &
  \multicolumn{2}{c|}{Geometric} &
  \multicolumn{2}{c|}{Var. Autoencoder} &
  \multicolumn{2}{c|}{StarGAN} \\ \cline{3-12} 
                        &        & Acc.  & F1    & Acc.  & F1    & Acc.  & F1    & Acc.  & F1    & Acc.           & F1             \\ \hline
\multirow{2}{*}{Gender} & Male   & 0.762 & 0.770 & 0.815 & 0.770 & 0.680 & 0.760 & 0.759 & 0.750 & \textbf{0.850} & \textbf{0.820} \\ \cline{2-12} 
                        & Female & 0.773 & 0.770 & 0.693 & 0.740 & 0.880 & 0.800 & 0.723 & 0.740 & \textbf{0.815} & \textbf{0.800} \\ \hline
\multirow{2}{*}{Age}    & Old    & 0.152 & 0.260 & 0.674 & 0.630 & 0.465 & 0.590 & 0.109 & 0.190 & \textbf{0.715} & \textbf{0.750} \\ \cline{2-12} 
                        & Young  & 0.978 & 0.690 & 0.533 & 0.570 & 0.885 & 0.730 & 0.960 & 0.670 & \textbf{0.774} & \textbf{0.780} \\ \hline
\multicolumn{12}{|l|}{* the models with the best performance (average accuracy and F1-score) are highlighted in bold}               \\ \hline
\end{tabular}%
}
\caption{Classification performance statistics of each model on the CelebA test set}
\label{tab:my-table}
\end{table}

According to the mean squared error and structural similarity index, the CelebA dataset is less similar to the UTKFace dataset compared to the LFWA+ dataset. Therefore, we would expect the overall performance to be quite low, as the model was trained on augmented versions of the UTKFace dataset. From a quick observation of table 5.4, we can confirm that this is the case. The gender and age accuracies of the best model tested on the UTKFace dataset is within the 85-95\% range, while the accuracies of the best model tested on the CelebA dataset ranges between 75-85\%. The performance also seems to be less balanced across the classes. Nevertheless, we can observe similar patterns in the performance, such as how training on images generated through variational autoencoders reduces the overall performance and consistency of the model. 

For the gender attribute, training a model on the StarGAN-generated images yields the best performance and consistency, with an overall accuracy of 83.3\% and standard deviation of 0.02 between the classes within the attribute. This is an improvement from the baseline accuracy of 0.768, although the baseline standard deviation is much lower at 0.008. However, the overall performance has dropped accuracy-wise when compared to the gender attribute classifier performance on the UTKFace dataset, which scored over 90\%. Other augmentation methods lead to a model that performs worse than the model and emphasizes the performance gap between the majority and minority classes. One interesting observation is that unlike what was observed from testing our model on the UTKFace and LFWA+ dataset, training on undersampled data actually increases the majority accuracy and reduces the minority accuracy for this particular attribute. Furthermore, we found that training on geometrically transformed images reduces the accuracy of the majority class while the accuracy of the minority class increases. This is a trend that was not present when running our classifier on the UTKFace and LFWA+ dataset.

Similarly, training on images generated by the StarGAN also yields the best performing model on the age attribute, although the accuracy is still quite low at an average of 74.5\%. However, since the baseline performance is very low with an accuracy of 56.5\%, this is considered as a substantial improvement. Furthermore, the performance disparity between the classes is quite significant in the baseline, and the best performing model was able to reduce the standard deviation between the classes in the attribute from 0.58 to 0.03. Therefore, in addition to improving the overall performance, training the model on the StarGAN-generated images also increases the model’s ability to make unbiased decisions. All of the other augmentation techniques aside from using variational autoencoders increase the performance of the minority class and reduce the gap between the performance of the majority and minority class.

From the above, we can conclude that even though there is a slight decline in performance, augmenting the dataset helps the generalization capability of the model on the CelebA dataset. Additionally, training on augmented data also consistently improves the performance of both the majority and minority class, as well as reducing the disparity between the performance of each class to a certain extent. Furthermore, we can also deduce that the similarity between an external dataset and the dataset a model is trained on affects the generalizability of the model on the external dataset.

\subsection{RQ3: Comparison with the FairFace Model}

In the previous section, we were able to identify which augmentation technique produces the best model. To get a general idea on how our model would compare in the object recognition field, we have made a comparison with the FairFace model. Due to the fact that the FairFace model was collected with an aim to have a dataset balanced across ethnicity, we would expect for it to perform the best when classifying the ethnicity attribute in the UTKFace and LFWA+ dataset. 

It is important to note that there are a few key differences between the model, namely the dataset it was trained on and the model that was trained. The FairFace model was trained on the FairFace dataset containing over 100,000 labeled images, while our model was trained on an augmented version of the UTKFace dataset. The FairFace model is based on the ResNet-34 architecture while our model is based on Inception v3. Although these factors might not allow us to make a direct comparison between the performance, we can still check how well our best model performs when compared to the state-of-the-art.

\begin{table}[h]
\centering
\resizebox{\textwidth}{!}{%
\begin{tabular}{|l|l|c|c|c|c|c|c|c|c|c|c|c|c|}
\hline
\multirow{3}{*}{Attribute} &
  \multirow{3}{*}{Class} &
  \multicolumn{4}{c|}{UTKFace} &
  \multicolumn{4}{c|}{LFWA+} &
  \multicolumn{4}{c|}{CelebA} \\ \cline{3-14} 
 &
   &
  \multicolumn{2}{c|}{Our model} &
  \multicolumn{2}{c|}{FairFace} &
  \multicolumn{2}{c|}{Our model} &
  \multicolumn{2}{c|}{FairFace} &
  \multicolumn{2}{c|}{Our model} &
  \multicolumn{2}{c|}{FairFace} \\ \cline{3-14} 
 &
   &
  Acc. &
  F1 &
  Acc. &
  F1 &
  Acc. &
  F1 &
  Acc. &
  F1 &
  Acc. &
  F1 &
  Acc. &
  F1 \\ \hline
\multirow{2}{*}{Gender} &
  Male &
  0.944 &
  0.910 &
  \textbf{0.951} &
  \textbf{0.950} &
  0.902 &
  0.905 &
  \textbf{0.954} &
  \textbf{0.980} &
  0.850 &
  0.820 &
  \textbf{0.969} &
  \textbf{0.970} \\ \cline{2-14} 
 &
  Female &
  0.891 &
  0.900 &
  \textbf{0.949} &
  \textbf{0.950} &
  0.917 &
  0.903 &
  \textbf{0.999} &
  \textbf{0.980} &
  0.815 &
  0.800 &
  \textbf{0.975} &
  \textbf{0.980} \\ \hline
\multirow{2}{*}{Age} &
  Old &
  \textbf{0.936} &
  \textbf{0.920} &
  0.670 &
  0.700 &
  \textbf{0.753} &
  \textbf{0.810} &
  0.572 &
  0.720 &
  \textbf{0.715} &
  \textbf{0.750} &
  0.139 &
  0.240 \\ \cline{2-14} 
 &
  Young &
  \textbf{0.890} &
  \textbf{0.910} &
  0.976 &
  0.970 &
  \textbf{0.890} &
  \textbf{0.830} &
  0.983 &
  0.820 &
  \textbf{0.774} &
  \textbf{0.780} &
  0.999 &
  0.890 \\ \hline
\multirow{4}{*}{Ethnicity} &
  White &
  \textbf{0.854} &
  \textbf{0.840} &
  0.952 &
  0.920 &
  \textbf{0.950} &
  \textbf{0.700} &
  0.967 &
  0.840 &
  - &
  - &
  - &
  - \\ \cline{2-14}  
 &
  Black &
  \textbf{0.886} &
  \textbf{0.900} &
  0.859 &
  0.900 &
  \textbf{0.790} &
  \textbf{0.810} &
  0.658 &
  0.730 &
  - &
  - &
  - &
  - \\ \cline{2-14} 
 &
  Asian &
  \textbf{0.892} &
  \textbf{0.890} &
  0.916 &
  0.900 &
  \textbf{0.650} &
  \textbf{0.700} &
  0.788 &
  0.260 &
  - &
  - &
  - &
  - \\ \cline{2-14} 
 &
  Indian &
  \textbf{0.854} &
  \textbf{0.860} &
  0.739 &
  0.790 &
  \textbf{0.575} &
  \textbf{0.540} &
  0.158 &
  0.640 &
  - &
  - &
  - &
  - \\ \hline
\multicolumn{14}{|l|}{* the models with the best performance (average accuracy and F1-score) are highlighted in bold} \\ \hline
\end{tabular}%
}
\caption{Classification performance comparison between our best model and the FairFace model}
\label{tab:my-table}
\end{table}

The results shown in table 5.5 are quite contradictory to our hypothesis. Given that the FairFace model was trained on data that was collected with the primary objective of having a dataset that was balanced ethnicity-wise, we would expect that our best models would perform consistently worse. Interestingly, the converse is true - our model outperforms ethnicity classification in both the UTKFace and LFWA+ dataset - with an overall accuracy of 80.6\%. Our model also exhibits a consistent performance between the different classes within the ethnicity attribute, with an overall accuracy standard deviation of 0.12. On the other hand, the FairFace model has an overall accuracy of 75.5\% and a standard deviation of 0.24. This shows that the most accurate model also tends to be the least biased.

Additionally, our model also consistently outperforms the FairFace model for age classification on all datasets, with an accuracy of 82.6\%. The overall accuracy of the FairFace model is quite low compared to ours, with an overall accuracy of 72.3\%. Our model’s performance is also more uniform across the various classes within each attribute, with an accuracy standard deviation of only 0.08, compared to the FairFace model’s standard deviation of 0.31. This shows that the FairFace model is more biased, and by looking at the table, it looks like the model constantly favors the majority class.

However, the gender classification performance by the FairFace model is unmatched across all the datasets, reaching accuracies which are constantly above 95\% on average, compared to our model’s 88.6\%. The FairFace model is also arguably more balanced with a standard deviation across the classes of 0.02, though our model is also fairly balanced with a standard deviation of 0.04. 

As said previously, we need to keep in mind that our models and the FairFace model have a different architecture, making the comparisons less meaningful. However, it is interesting to see how our model performs against the state-of-the-art annotator with respect to classification accuracy, generalizability across various datasets, and bias mitigation.


\chapter{Conclusions}

In this project, we have successfully evaluated the performance of several methods on alleviating the effect of dataset biases on the final model, primarily caused by imbalances within classes. We have implemented the various data augmentation techniques and trained a multi-class classification model on the augmented data. From our thorough experimentation and analysis of the result, as well as comparison with existing state-of-the-art models, we were able to show how some of our techniques can aid the dataset imbalance problem and create an unbiased model that matches the performance of current state-of-the-art.

In summary, we have taken the UTKFace dataset and extracted the gender, age, and ethnicity attributes for our experiments. We apply each balancing technique - undersampling, geometric transformations, variational autoencoders and generative adversarial networks - on the training data and train a classifier to classify the different attributes on the held-out test set. We evaluated their performance by analyzing the per-class accuracy and F1-score and identifying any inconsistencies. We have also performed cross-dataset evaluation to understand the generalizability performance of our model on external datasets with the same domain. Furthermore, we also found that there is a positive correlation between the generalizability of the model on an external dataset and the similarity between the external dataset and the dataset that the model was trained on. Finally, we have discussed the key observations from the various experiments and critically evaluate each balancing method and data augmentation technique on their capability to be used to obtain a fair and unbiased model.

Through rigorous experimentation, we were able to show that some of our data augmentation techniques were able to mitigate the biases in the model caused by class imbalances within the training dataset. From the dataset, we were then able to train a multi-attribute classifier with a better generalization performance than our initial baseline. It is also important to note that our classifier performs consistently better on the binary age attribute than the state-of-the-art ResNet-based attribute classifier trained on the FairFace dataset. One key observation is that although most of the best models were obtained by training a model on a dataset containing StarGAN-generated images, the models trained on geometrically-transformed images do not fare too far behind. However, geometrically transforming images and adding them to the training set do not take as much time as training a StarGAN and then generating images from the model. Therefore, keeping in mind the trade-off between accuracy and training speed, we found that training a model on a dataset augmented through geometric transformations can provide you with an unbiased high-performing classifier in a short amount of time.

\bibliographystyle{plain}
\bibliography{mybibfile}

%
%
%

\end{document}